\documentclass[runningheads]{llncs}
\usepackage{multirow} 
\usepackage{graphicx}
\usepackage{booktabs}
\usepackage{xcolor, colortbl}
\definecolor{almond}{RGB}{186,210,225}

\usepackage{subcaption}
\begin{document}
\title{Knowledge-grounded Adaptation Strategy for Vision-language Models: Building Unique Case-set for Screening Mammograms for Residents Training}
%
%\titlerunning{Abbreviated paper title}
% If the paper title is too long for the running head, you can set
% an abbreviated paper title here
%

\author{Paper ID 3702}
\authorrunning{}
% First names are abbreviated in the running head.
% If there are more than two authors, 'et al.' is used.
%
\institute{
% Department of Radiology, Mayo Clinic \and
% Department of Radiology, UW Madison School of Medicine and Public Health \and
% School of Computing and Augmented Intelligence, Arizona State University
}

\author{Aisha Urooj Khan\inst{1} \and
John Garrett \inst{2} \and
Tyler Bradshaw\inst{2} \and
Lonie Salkowski\inst{2} \and
Jiwoong Jason Jeong\inst{3} \and
Amara Tariq \inst{1} \and
Imon Banerjee\inst{1,3}}
\authorrunning{A. Urooj et al.}
% First names are abbreviated in the running head.
% If there are more than two authors, 'et al.' is used.
%
\institute{
Department of Radiology, Mayo Clinic \and
Department of Radiology, UW Madison School of Medicine and Public Health \and
School of Computing and Augmented Intelligence, Arizona State University
}

% }
%
\maketitle              % typeset the header of the contribution
\begin{abstract}
A visual-language model (VLM) pre-trained on natural images and text pairs poses a significant barrier when applied to medical contexts due to domain shift. Yet, adapting or fine-tuning these VLMs for medical use presents considerable hurdles, including domain misalignment, limited access to extensive datasets, and high class imbalances. Hence, there is a pressing need for strategies to effectively adapt these VLMs to the medical domain, as such adaptations would prove immensely valuable in healthcare applications. In this study, we propose a framework designed to adeptly tailor VLMs to the medical domain, employing selective sampling and hard-negative mining techniques for enhanced performance in retrieval tasks. 
We validate the efficacy of our proposed approach by implementing it across two distinct VLMs: the in-domain VLM (MedCLIP) and out-of-domain VLMs (ALBEF).
We assess the performance of these models both in their original off-the-shelf state and after undergoing our proposed training strategies, using two extensive datasets containing mammograms and their corresponding reports. Our evaluation spans zero-shot, few-shot, and supervised scenarios. Through our approach, we observe a notable enhancement in Recall@K performance for image-text retrieval task.
\end{abstract}

\section{Introduction}

According to the American Cancer Society (ACS) screening guidelines, women between 40 and 44 have the option to start screening with a mammogram every year and women 
45 to 54 should get mammograms every year. This resulted a huge number of screening mammogram exams at each healthcare institution and consumes significant radiologists time for reading. One study showed a 40\% disparity among radiologist screening sensitivity and a 45\% range in the rates at which women without breast cancer are recommended for biopsy~\cite{beam1996variability}. 

\begin{figure}[tb!]
  \centering
\begin{subfigure}{0.32\textwidth}
    \includegraphics[width=\textwidth]{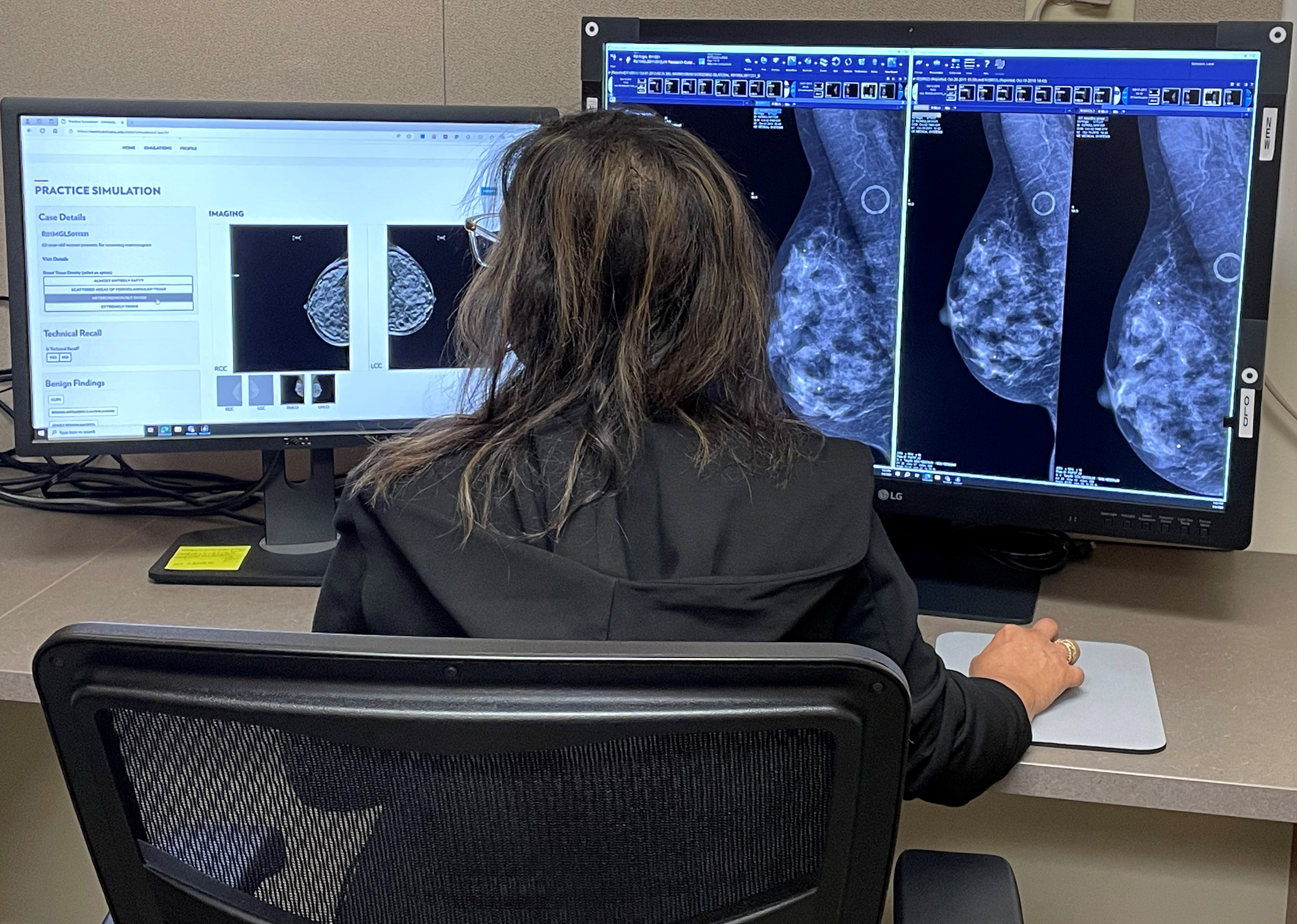} 
    \caption{}
\end{subfigure}
\begin{subfigure}{0.52\textwidth}
    \includegraphics[width=\textwidth]{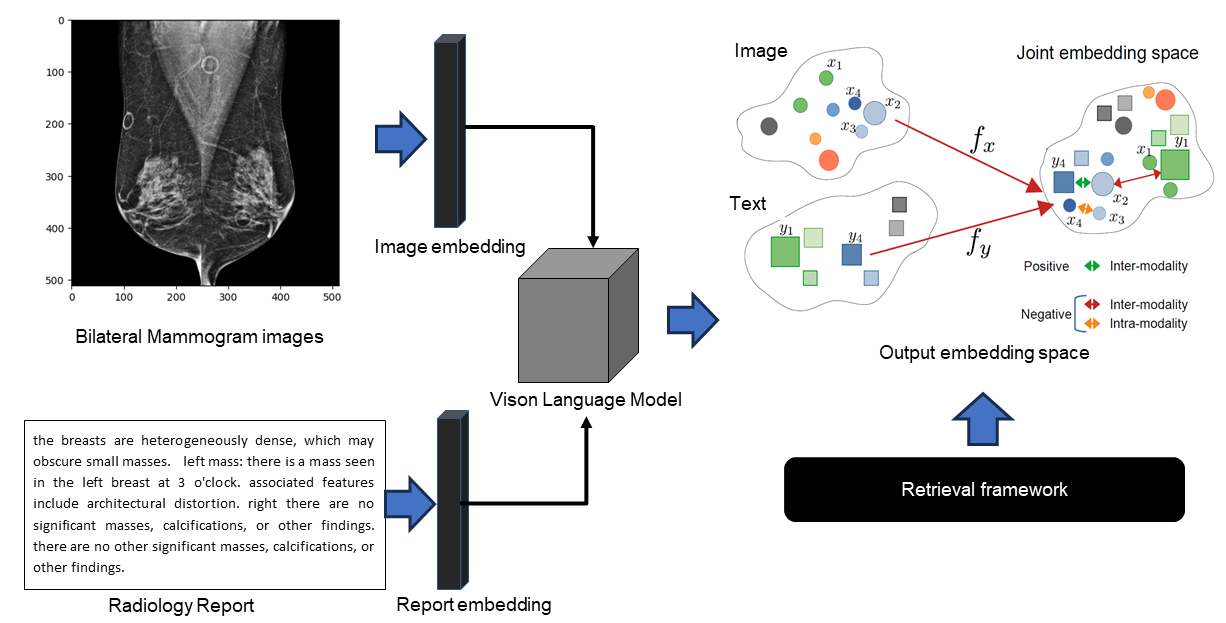}
    \caption{}
\end{subfigure}
   \vspace{-15pt}
\caption{\scriptsize Multimodal learning for screening mammogram: (a) a session with radiology resident for the case review; (b) framework generating joint embedding space for bilateral mammogram and free-text radiology reports. Illustration of joint embedding space (right) is adapted from CrossCLR~\cite{zolfaghari2021crossclr}.}
   \label{fig:framework}
    \vspace{-15pt}
\end{figure}
During 12 weeks of required residency training in breast imaging, the Accreditation Council for Graduate Medical Education (ACGME) requires residents to document a minimum of 300 interpretations of breast imaging exams (mammograms, ultrasounds, MRI) and there is no particular criteria for training case-selection~\cite{davis2006accreditation}. Even after this requirement, the majority (59\%) of residents do not feel prepared to read mammograms after completing their training~\cite{bassett2003survey}. Unfortunately, the number of fellowship-trained breast imaging radiologists is expected to decline and thus the majority of residents will face reading mammography as part of their eventual clinical practice. 
The fundamental fear of misdiagnosis (missing a cancer) and the feeling that residency does not fully prepare them to read mammograms, likely contributes to an increase in additional mammogram scans to confirm diagnosis and incur avoidable cost and effort~\cite{miglioretti2009radiologists}.
Thus, providing adequate training with relevant case-selection within radiology residency will benefit more women and bestow safer mammographic interpretation.  
However, hand picking a set of such relevant cases is both time-consuming and challenging, as well as can introduce sampling bias and is unlikely to match the desired distribution. 
% Furthermore, most PACS system that have search tools with very limited search criteria which often resulted countless useless cases. 
Deep learning retrieval framework has the potential to automate and optimize case selection from 100,000’s of cases based on multimodal data - imaging features and textual findings documented within the reports. \\
We develop a multimodal framework 
% based on vision language model (VLM)
to automatize the relevant case-selection based on both text and image representation of the individual screening exams (Fig~\ref{fig:framework}). However, there are inherent technical challenges for training such a model - (i) natural image pre-trained VLM often unable to capture the radiology vocabulary with selective terms and also natural image features does not corresponds well with gray-scale and small mammograpy findings; (ii) relevant abnormal imaging findings (mass, calcification, architectural distortion, solitary dilated duct) are rare in screening mammogram which makes the model primarily learn the negative cases and omit the actual findings; (iii) syntactic difference between the semi-structured reports are minimal, and thus the reports with very different findings resulted similar embeddings; (iv) variations in breast density is often the most prominent image feature in mammogram and high density can occlude abnormal imaging features. To deal with the above mentioned challenges, we propose a \emph{knowledge-based grouping of the mammogram cases}, \emph{selective sampling, and hard-negative mining techniques for VLM model training}. We validate the efficacy of our proposed approach across two distinct VLMs: the in-domain VLM (MedCLIP) and out-of-domain VLM (ALBEF).
% We assess the performance of these models both in their original off-the-shelf state and after undergoing our proposed training strategies, using Institute X datasets containing mammograms and their corresponding reports.
Our evaluation spans zero-shot, few-shot, and supervised scenarios using Institute X datasets containing mammograms and their corresponding reports. The model was also externally validated on screening mammogram data from Institute Y.
% with significant shift in population distribution. 

\section{Methodology} \label{sec:method}
\vspace{-5pt}

Given a vision-language model $f(\theta)$, we want to train $f(\theta)$ effectively such that similar image-text pairs $(I_p, T_p)$ are close to each other in semantic space. 
% These VLMs are trained using a loss function to align paired image and report closer and push negative image/report farther.
Negative pairs are often picked within a batch from a different data sample. 
% Applying this strategy directly for radiology image-report pairs may not yield the optimal results. 
% The radiology reports usually follow a structure/template unlike captions in natural image-text datasets. 
% Given an image-report pair$(I_p, T_p)$ for the mammogram case, the report comprises of radiology observations made on the examination image. 
For any given medical sub-domain, the vocabulary to describe the observations largely stays consistent, particularly in mammogram as the reports are formulated following the standardized BIRADS vocabulary~\cite{lazarus2006bi} generated by  the American College of Radiology (ACR). 
% for standardized the assessment and reporting of breast lesions identified on mammograms. 
These image-report pairs can be grouped based on the important findings in a way that each image-report pair with same concepts belong to one group. 
% An easy case is a normal exams with no mention of specific image features representative of abnormality, will be considered a `no finding'. 
\begin{figure*}[t]
\footnotesize
   \centering

   \includegraphics[width=0.9\textwidth]{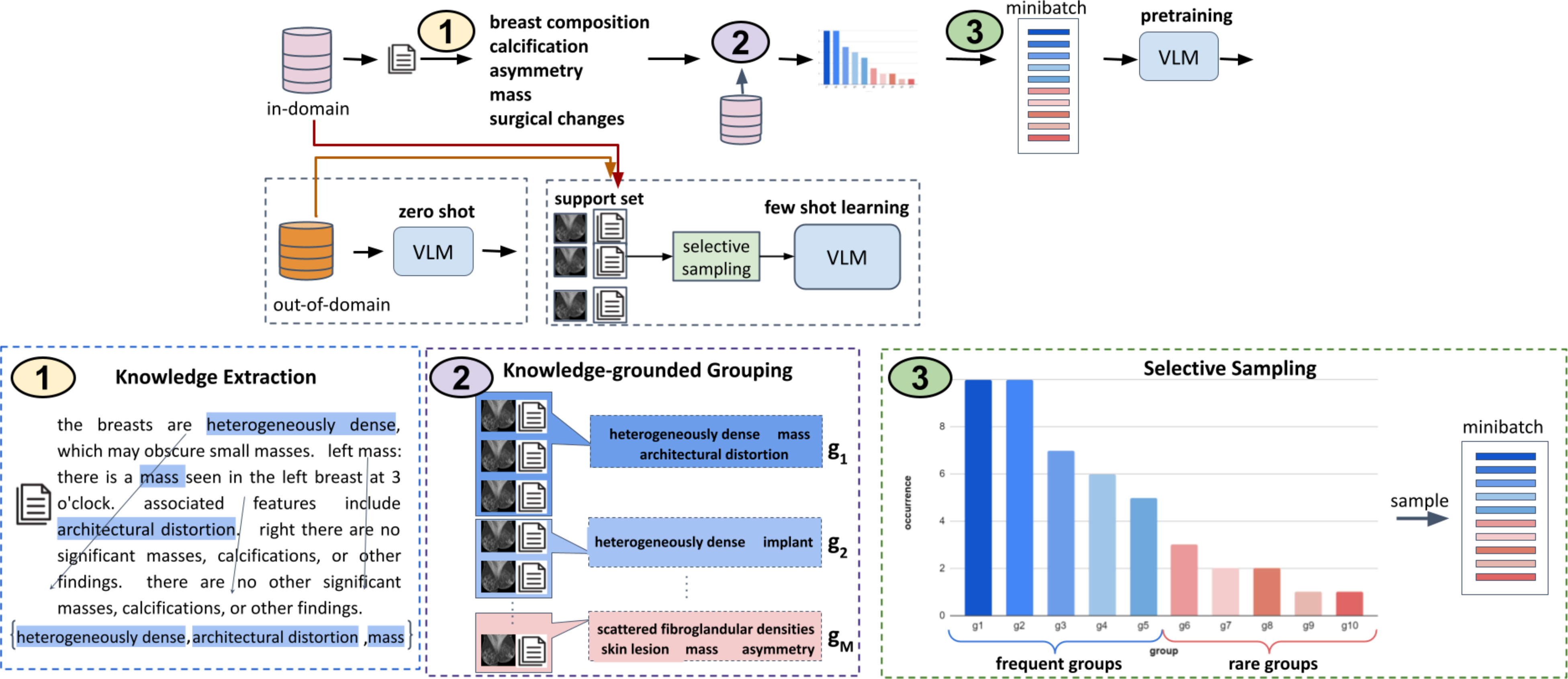}
   \caption{\footnotesize Workflow for adapting the VLM with the proposed
   % 1) In-domain knowledge is extracted from reports to obtain the unique combinations (groups) of BIRADS image descriptors present in text. 2) These groups are utilized for knowledge-grounded grouping of image-text pairs. 3) Perform selective sampling over the grouped image-report pairs such that pairs from unique groups are chosen in the mini-batch ensuring a certain ratio of rare groups is also present within minibatch. Model is pretrained using
   selective sampling to learn joint representation aware of fine-grained knowledge. The pretrained model is tested on out-of-domain data for zero shot evaluation. For few shot learning, support set is obtained from the training data to fine-tune model.
   }
   \label{fig:main}
     \vspace{-15pt}
\end{figure*}
% In traditional constrastive learning, a negative pair is considered any mismatch in image or text except the true image-text pair. 
Additionally, for mammograms,
% reports are often template-based, and 
broad features are visually similar to each other and need a domain expert, i.e., a radiologist to examine for anomalies. 
% Additionally, $\sim90\%$ of the images are considered normal (`no finding') cases, the anomalies found in the remaining $\sim10\%$ images also have an additional class imbalance on top of the high disparity in normal and abnormal cases. 
Given the textual and visual similarity between the cases, there is a high chance that the sampled `negative' image $I_n$ or text report $T_n$ has the similar findings as the true pair does. This leads to confusing the model during training because it might be pushing away semantically similar image-text pairs.
% that have actual similar semantics. 
We propose to sample a mini-batch in a way that within batch negatives are ensured to be coming from true negatives and minority cases are equally represented during training. This is achieved in three steps as described below: 
% 1) knowledge extraction from image-text pairs, 2) knowledge-grounded grouping them based on the unique combination of concepts found in step 1, and lastly, 3) selective sampling of mini-batches during training. 
\vspace{-5pt}
\paragraph{\textbf{1)Knowledge extraction:}} \label{sec:knowledge_extract}
% \vspace{-5pt}
 To form the groups, we leveraged the standard 54 unique BIRADS image descriptors and extracted the positive mentioned from the radiology reports which are lower cased and cleaned before extracting key concepts. For example, for the following text report:``
   \emph{ the breasts are \textcolor{blue}{heterogeneously dense}, which may obscure small masses.   left mass: there is a \textcolor{blue}{mass} seen in the left breast at 3 o'clock. associated features include \textcolor{blue}{architectural distortion}.   right there are no significant masses, calcifications, or other findings''},
the  extracted group is $\{$heterogeneously dense, mass, architectural distortion$\}$ based on the key concepts highlighted in blue. In the context of mammograms, the abnormal image descriptors are primarily categorized into 10 groups - breast composition, calcification, asymmetry, mass, surgical changes. All of these concepts except tissue density may or may not be present in the normal image without anomaly. We excluded all the negative and uncertain findings.  
\vspace{-5pt}
\paragraph{\textbf{2) Knowledge grounded grouping:}}
\vspace{-5pt}
% We leverage the grouped image-text pairs based on the 54 key observations defined in the BIRADS vocabulary such as tissue density, features, surgical changes, etc.

The presence of a key concept combination in any exam is considered a group such that every other image with the same key concepts present belongs to the same group. All text reports with the same key concepts (even ordered differently - $<A,B,C>$ vs $<B,C,A>$) belong to the same group. This yields a unique set of groups from the extracted knowledge for the given dataset. 
Formally, a group $g_i\in G^{M}$ for $i\in1,2,...,M$ is a set of key concepts within an image extracted from the paired radiology report, where $G^{M}$ is the set of $M$ total groups extracted from the text reports.
% Negatives are defined as image-text pair belongs to a different group while some features may be common between them, such as positive group $<A,B>$ vs negative group $<A, B, C>$. 
% Such knowledge grounded grouping helps to define `hard negatives' with certain importance for feature combinations to properly represents the complexity of the cases. 
\vspace{-5pt}
\paragraph{\textbf{3) Selective Sampling:}} \label{sec:selective_sampling}
\vspace{-5pt}
Given an image $I_p$ and paired text report $T_p$ as $(I_p, T_p)$, a negative pair is denoted by $(I_p, T_n)$ or $(I_n, T_p)$, where $I_n$ and $T_n$ belong to an instance from a different group. 
% Once we know the set of groups $G^{M}$, 
For each pair $(I_{p_i}, T_{p_i})$ from group $g_i$, a negative image $I_{n_j}$ or text $T_{n_j}$ can be selected from group $g_j\in G^M$ when $j \ne i$. This approach while addresses the challenge of alike image-text pairs within a mini-batch, it still faces the long-tail distribution challenge due to class imbalance. As frequent groups have a high chance of being sampled, rare groups often might never be seen during training. To address this problem, a mini-batch is sampled based on the group frequency. 
% More specifically, 
We define a heuristic-based boundary $b$ to sample rare groups such that $b < batch\_size$ and  $batch\_size - b$ instances are selected from groups with high occurrence, i.e., frequent groups. This ensures that $b$ instances are coming from rare groups, where rare and frequent groups are empirically chosen based on the data distribution.
\paragraph{\textbf{VLM Training}}
\vspace{-5pt}
The proposed sampling strategy can be used to sample mini-batches to train the vision-language model for contrastive learning. We use sampling strategy in two settings: pretraining and few-shot learning across two existing VLMs: ALBEF~\cite{ALBEF} and MedCLIP~\cite{medclip}.
% ALBEF and MedCLIP are two-stream VLMs whereas ViLT has a single encoder for images and text. 
\emph{Evaluation Metrics:}
% On the joint embedding space, we measure the retrieval performance separately on both text and image query. 
To measure the performance, we consider the Recall@K metric and report top-1, top-5, and top-10 performance. 
% Because of the groups we construct by preprocessing report findings,
We consider it a success if any report with the same findings (hence the same group) appears in the top-K ranks. 
% For few-shot experiments, we consider K=10 shots to finetune the models.

%----------------------------------------------------------------------
\vspace{-5pt}
\section{Experiments and Results}
\vspace{-5pt}
\paragraph{\textbf{Datasets:}}
\emph{Internal Dataset:} Using IRB approval, we collected 72,328 bilateral screening mammogram exams from 46,848 patients acquired between January 2016  December 2018 from Institute X health affiliated centers as our internal dataset. We randomly split the dataset into train-val-test with 70,238 $<image-report>$ pairs used for training, 1000 image-report pairs for validation, and 1000 image-report pairs as a test set respectively.
% to report the model's performance. 
% The internal dataset's population of 58.7 average age (median age: 59, interquartile range: 15 [51, 66]) includes ~92\% white, ~3\% black, 2.5\% Asian, and the remaining 2.5\% are other/unknown. 
% The exams contains digital breast tomosynthesis (DBT) combined with digital mammography, and we selected Left-MLO and Right-MLO 2D view from the digital mammography. 
We use a binary mask of thresholded pixel values to identify the largest connected component in the image and use its bounding box coordinates to crop the breast tissue area. The cropped R-MLO and L-MLO images are concatenated, zero-padded for maintaining the aspect ratio, and resized to $512\times512$ pixels.
Reports are cleaned by lowercasing, punctuation removal, and extra spacing removal. The text is then split into sentences, each examined for key concepts: density, calcifications, asymmetry, architectural distortion, mass, and additional features. Negation sentences are ignored. If a sentence contains a key concept, the report is marked accordingly. Each key concept is detected separately and then combined to form discrete groups.
% The breast tissue area is cropped from the individual images, and pixel data from both views are then stitched together, originating from the chest wall. The stitched images are zero padded to make them a square image, then resized to $512\times512$. 
% Findings from semi-structured reports are paired with stitched images after cleaning. More specifically,
% Reports are processed to extract and group the BI-RADS concepts (relevant for breast cancer screening) from the finding sections, e.g., `mass', `tissue density', `surgical changes', and `architectural distortion' are extracted from each report. 
% Because, these reports are templated instead of free text, reports with the presence of certain concepts are syntactically same. 
This grouping allows selective sampling during model training as described in \ref{sec:selective_sampling}. We find 1005 unique groups in the train set. 
% Group statistics for the internal test set w.r.t preprocessing of findings are shown in fig~\ref{fig:test_data_dist}.
Detailed group distribution is provided in the supplementary document.
\emph{External Dataset:} With the Institute Y IRB approval, the screening mammogram collected between 2018 - 2022 is used for external validation of our approach for supervised training as well as few shot learning. 
% From the Institute Y dataset, we consider screening image-report pairs processed the similar way as internal dataset. 
% There are 9,187 image-report pairs in
Institute Y dataset has
% is randomly split in 80\%-20\% train-test splits with
8,172 training image-report pairs and 1,015 pairs in test set. The test set is then used for external validation. The test set has 79 unique groups after preprocessing as described in section~\ref{sec:knowledge_extract}.
% Unlike Institute X templated reports, reports in Institute Y data are in free-style text format is also added additional challenges during the external validation of VLMs .

%----------------------------------------------------------------------

%----------------------------------------------------------------------

%----------------------------------------------------------------------

\vspace{-5pt}

\paragraph{\textbf{Implementation Details:}}
% \paragraph{\textbf{ALBEF}~\cite{ALBEF}:}
ALBEF~\cite{ALBEF} is a VLM
% has an image encoder and a text encoder, followed by a cross encoder
with image-text contrastive loss. 
% An image-text alignment loss is used to align image and text features even before cross-attention. Image-Text Matching (ITM) and Masked LM (MLM) are used in addition to jointly optimize the model. 
% The baseline ALBEF model initializes from DeiT~\cite{deit} vision transformer using $16\times16$ patch size. 
We pretrain ALBEF on Institute X image-report pairs, followed by a retrieval-only task Image Text Matching (ITM) for fine-tuning the pretrained backbone named ALBEF-Ret. For a $512\times512$ image and the patch size of $16\times16$, image encoder takes $1024$ patch tokens 
% ($512\div16=32;32\times32=1024$)
in the ALBEF model. We train ALBEF with (ALBEF-SS) and without (ALBEF-Ret) the proposed selective sampling. 
% We validated the model on the Institute Y external dataset. 
% \paragraph{\textbf{MedCLIP}~\cite{medclip}:} 
% \textbf{MedCLIP}~\cite{medclip} is a variant of CLIP~\cite{radford2021learning}  model
% is originally 
% trained on gray-scale chest x-rays and reports. 
We evaluate MedCLIP~\cite{medclip} pretrained on CheXPERT dataset~\cite{irvin2019chexpert}  and MIMIC-CXR~\cite{johnson2019mimic} for zero-shot, initialize model weights for few shot learning, and train MedCLIP on the 2D mammogram images for fully supervised backbone. Similar to ALBEF, we also trained MedCLIP with (MedCLIP-SS) and  
without (MedCLIP) the proposed selective sampling. 
% We used the same internal and external validation sets.
For full training, we consider top 20 groups w.r.t the number of samples as frequent groups out of total 1005 unique groups. 
% Ratio of frequent to rare groups in a minibatch is set to R=0.375.
We use batch size=8 and boundary b=3 for random sampling of frequent and rare groups, i.e., for R=0.375 - 5 instances belong to frequent groups, and 3 are sampled from the set of rare groups.
All training parameters except the hyperparameters considered for this study stay the same across models. 

% \subsubsection{ViLT~\cite{vilt}:} ViLT is a single multimodal encoder based model, a ViT~\cite{} like model trained on concatenated image patch tokens and paired text tokens. We fine-tune the model using their vanilla approach and compare with ViLT-SS trained with the proposed selective sampling. 

\begin{table*}[t]
\tiny
\renewcommand{\arraystretch}{.9}
  \centering \setlength{\tabcolsep}{.9\tabcolsep}   
    \begin{tabular}{ll|cccc|ccc}
       \toprule
                % \multicolumn{1}{c}{\textbf{Method}} & \multicolumn{3}{c}{}  \\
                % \cmidrule(lr){2-4}

    & & \multicolumn{3}{c}{Internel test set} & & \multicolumn{3}{c}{Externel test set}  \\
   \cmidrule{3-5}  \cmidrule{7-9} 
 \textbf{Task} & Model & R$@$1 & R$@$5  & R$@$10  && R$@$1  & R$@$5  & R$@$10 \\
      
        \midrule

\multirow{5}{*}{Image-to-Report}    & NN(k=10) & 10.1 & - & - && 3.34 & - & - \\
    % & Random  &  &  &  &&  &  & \\
    \cmidrule{2-9}
   & ALBEF-Ret  & 12.9 & 37.0 & 47.2 && 19.00 & 50.21 & 65.76\\

     & ALBEF-SS-PT (ours) & 9.0 & 32.3 & 40.2  && 20.25 & 48.75 & 51.56\\

   &\cellcolor{almond} ALBEF-SS-Ret (ours) &\cellcolor{almond} \textbf{30.5} & \cellcolor{almond}\textbf{53.9} &\cellcolor{almond} \textbf{61.3} && \cellcolor{almond}\textbf{21.61} & \cellcolor{almond}46.03   & \cellcolor{almond}55.22 \\
\cmidrule{2-9}
    & MedCLIP & 6.4 & 11.2 & 15.1 && 16.6 & 30.27 &  35.17\\
          &\cellcolor{almond} MedCLIP-SS (ours) & \cellcolor{almond}5.10 & \cellcolor{almond}10.60 &\cellcolor{almond} 14.90 && \cellcolor{almond}4.28 & \cellcolor{almond}11.69 &  \cellcolor{almond}20.98\\
      % &\cellcolor{almond} MedCLIP-SS (ours) & \cellcolor{almond}1.1 & \cellcolor{almond}5.1 &\cellcolor{almond} 9.9 && \cellcolor{almond}\textcolor{red}{8.98} & \cellcolor{almond}\textcolor{red}{39.04} &  \cellcolor{almond}\textcolor{red}{45.93}\\
    % & MedCLIP-MC & 0.64 & 11.2 & 15.1 &&  &  & \\
    %   & MedCLIP-SS-MC & 1.1 & 5.1 & 9.9 &&  & &  \\
    % & MedCLIP-SS-B16 (ours) & 1.5 & 6.0 & 12.0 && 2.5 & 24.84 & 42.91\\

     % & VilT &  & - &  & &  &  &  \\

     % & VilT-SS-PT (ours) &  & - &  & & &  & \\
   
    \midrule
    \midrule
    
     \multirow{5}{*}{Report-to-Image}  & NN(k=10) & 26.4 & - & - && 36.95 & - & -\\
 
  % & Random  &  &  &  &&  &  & \\
  \cmidrule{2-9}
     & ALBEF-Ret  & 28.6 & 60.5 & 65.2 && 34.13 & \textbf{82.98} & 83.82 \\

     & ALBEF-SS-PT (ours) & 19.4 & 60.7 &  67.6 &&  \textbf{63.88} & 81.73 & 84.76\\

   &\cellcolor{almond} ALBEF-SS-Ret (ours) & \cellcolor{almond}\textbf{35.8} & \cellcolor{almond}\textbf{63.3} & \cellcolor{almond}\textbf{73.4} && \cellcolor{almond}54.70 & \cellcolor{almond}81.94 &\cellcolor{almond} \textbf{85.49}\\
\cmidrule{2-9}
    & MedCLIP & 26.70 & 48.40 & 56.30 && 0.31 & 20.77 & 22.02 \\

     & \cellcolor{almond}MedCLIP-SS (ours) & \cellcolor{almond} \textbf{31.5}&\cellcolor{almond}\textbf{62.3 } & \cellcolor{almond} \textbf{66.2} && \cellcolor{almond}\textbf{ 0.52} & \cellcolor{almond} \textbf{21.4}& \cellcolor{almond}\textbf{ 24.22}\\
    
      % & \cellcolor{almond}MedCLIP-SS (ours) & \cellcolor{almond}23.8 &\cellcolor{almond} 73.8 & \cellcolor{almond}78.1 && \cellcolor{almond}\textcolor{red}{17.64} & \cellcolor{almond}\textcolor{red}{21.71} & \cellcolor{almond} \textcolor{red}{44.15}\\
      
      %    & MedCLIP-MC & 26.7 & 48.4 & 56.3 &&  &  & \\
      % & MedCLIP-SS-MC & 23.8 & 73.8 & 78.1 &&  & &  \\
      
         % & MedCLIP-SS-B16 (ours) & 15.4 & 47.2 & 55.7 && 37.89 & 48.85 & 54.28 \\
      % & MedCLIP-SS (B=48) (ours) &  &  &  &&  &  &  \\
    % & MedCLIP-SS-Ret (ours) & &  && &  &  & \\

     % & VilT &  & - &  & &  &  &  \\

     % & VilT-PT (ours) &  & - &  & & &  & \\

\bottomrule

    \end{tabular}
 \vspace{5pt}
    \caption{\footnotesize Comparative retrieval results for the proposed knowledge grounded selective sampling (SS) on both internal (Institute Y) and external (Institute Y) test sets. `Ret':fine-tune models, `PT':pre-trained model. Numbers are in percentages.}
    \label{tab:vlm}
    \vspace{-7mm}
% \vspace{-15pt}
\end{table*}

% \subsubsection{Results:}
\noindent \textbf{\emph{Results:  Image$\leftrightarrow$Text retrieval:}}
We evaluate the learned joint embedding using image-text retrieval (ITR) as our downstream task. Here, we compare ALBEF with ALBEF-SS, and MedCLIP with MedCLIP-SS to assess the impact of selective sampling during training. We observe improvement for both VLMs for image-to-report and report-to-image retrieval, and discuss performance on our internal test set as well as external test data. Table~\ref{tab:vlm} presents the complete results on the internal and external data. 
% \textbf{\textit{Internal test set:}}
More specifically, on internal test set, ALBEF-SS-Ret obtains 17.6\% $\uparrow$ gain in R@1 performance, $\sim$17\%$\uparrow$ improvement in R@5, and 14.1\%$\uparrow$ increase in R@10 score over ALBEF-Ret model for image-to-report retrieval. For report-to-image retrieval, ALBEF-SS-Ret improves by 7.2\% $\uparrow$ at R@1, 2.8\% $\uparrow$ at R@5, and 8.2\% $\uparrow$ at R@10 scores. 
% The baseline MedCLIP achieves R@1:6.4\% , R@5:11.2\% and R@10:15.1\% for image-to-report. 
MedCLIP-SS achieves comparable results to the MedCLIP baseline for R@5 and R@10. For report-to-image retrieval, MedCLIP-SS achieves performance gain of 4.8\%$\uparrow$ in R@1, 1.8\%$\uparrow$ as R@5, and with a significant margin of $\sim$10\%$\uparrow$ in R@10 respectively. Overall, we observe that image-to-report retrieval is more challenging task for VLMs compared to report-to-image retrieval. 
% \textbf{\textit{External test set:}} 
On external test set, 
% for image-to-report retrieval, the trend is however flipped compared to internal set.
ALBEF-SS-Ret model although improves over ALBEF by 2.61\% in terms of R@1, but performance is hurt on R@5 and R@10. Similar behavior is observed for MedCLIP-SS as well.
% R@5, and R@10 by 21.31\% $\uparrow$, 16.07\%$\uparrow$, and a slight gain of 0.21\% $\uparrow$ respectively.
However, we notice consistently significant improvement in both ALBEF-SS-Ret and MedCLIP-SS for report-to-image retrieval. 
% ALBEF-SS-Ret, compared to its vanilla variant,  obtains a huge gain of 20.57\% $\uparrow$ in R1, and achieves R10 score of 85.49\%  improving by $\sim$1.67\% $\uparrow$. With selective sampling, the pretrained version ALBEF-SS-PT obtains 63.88 R1 score ($\sim$30.0\% $\uparrow$) and a slight improvement for R10. 
MedCLIP-SS consistently performs better than MedCLIP in terms of R@1, R@5, and R@10 respectively. 
% on the other hand, we observe an increase in R5 and R10 for both image-to-report and report-to-image retrieval despite decrease or comparable performance in R1 scores. 

% Fig.~\ref{fig:before-vs-after} presents the case for vanilla ALBEF model when image-text alignment loss did not follow the patterns of other loss objectives used in training the model, and after using the selective sampling, the alignment loss pattern changed completely. 

\begin{table*}[t]
\tiny
\renewcommand{\arraystretch}{.9}
  \centering \setlength{\tabcolsep}{.9\tabcolsep}   
    \begin{tabular}{lll|cccc|ccc}
       \toprule
                % \multicolumn{1}{c}{\textbf{Method}} & \multicolumn{3}{c}{}  \\
                % \cmidrule(lr){2-4}

     & &  &\multicolumn{3}{c}{Internal test set} & & \multicolumn{3}{c}{External test set}  \\
   \cmidrule{4-6}  \cmidrule{8-10} 
 \textbf{Task} & K & Model  & R$@$1 & R$@$5  & R$@$10  && R$@$1  & R$@$5  & R$@$10 \\
      
        \midrule

% \multirow{8}{*}{Image-to-Report}    & \multirow{3}{*}{1} & MedCLIP & 2.3 & 10.9 &  24.3 && 34.86 & 55.95 & 64.92 \\
%  % &  & MedCLIP-PT (ours) & 6.9 & 18.2 & 26.3 &&  &  &  \\
%   &  & ALBEF &  &  &  &&  &  &  \\
%     &  & ALBEF-PT (ours) & 19.6 & 41.1 & 50.9 &&  &  &  \\

\multirow{8}{*}{Image-to-Report}    & \multirow{4}{*}{ZS} & MedCLIP-ViT & 1.9 & 12.0 & 20.5 && \textbf{25.71} & 38.42 & 40.79 \\
& & ALBEF-mscoco  & 16.8 & 32.0 & 40.5 && 14.61 & 36.01 & 43.11\\
   &   & ALBEF-flickr30k & 20.0 & 31.1 & 37.5  && 7.83 & 33.82 & 40.29\\
  &   & \cellcolor{almond}ALBEF-SS-Ret (ours) &\cellcolor{almond}- & \cellcolor{almond}- & \cellcolor{almond}-  && \cellcolor{almond}21.61 & \cellcolor{almond}\textbf{46.03 }  &\cellcolor{almond}\textbf{ 55.22}\\
    \cmidrule{2-10}  
   % & \multirow{3}{*}{5} & MedClip & 6.5 & 34.5 & 56.1 && 22.17  & 71.92 & 77.93 \\
   %  &  & MedClip-upto K &  &  &  && 28.47 & 65.81 & 76.16 \\
   %   &  & ALBEF &  &  &  &&  &  &  \\
   %  &  & ALBEF-PT (ours) &  &  &  &&  &  &  \\
   %  \cmidrule{2-10}
& \multirow{3}{*}{10} & MedCLIP & 0.1 & 3.1 & 6.8 && \textbf{32.36} & \textbf{48.43} & \textbf{57.09} \\
   % & \multirow{3}{*}{10} & MedCLIP-old & 4.3 & 16.2 & 26.8 && 5.42 & 50.00 & 61.48 \\
   & &\cellcolor{almond} MedCLIP-SS &\cellcolor{almond} \textbf{2.2} & \cellcolor{almond}\textbf{8.0} &\cellcolor{almond} \textbf{14.1} && \cellcolor{almond}18.00 & \cellcolor{almond}36.22 &\cellcolor{almond} 41.44 \\
  \cmidrule{3-10}
     % &  &\cellcolor{almond} MedCLIP-SS-g20 (ours) &\cellcolor{almond} 3.2 & \cellcolor{almond}12.7 &\cellcolor{almond} 23.1 && \cellcolor{almond}7.8 & \cellcolor{almond}31.63 & \cellcolor{almond} 48.54 \\
     
     &  & ALBEF & 19.5 & 46.9 & 55.0 && 0.3 & 29.96 & 55.01 \\
      &  & \cellcolor{almond}ALBEF-SS-Ret & \cellcolor{almond}\textbf{25.40} & \cellcolor{almond}\textbf{48.10} & \cellcolor{almond}\textbf{57.40} && \cellcolor{almond}\textbf{20.88} & \cellcolor{almond}\textbf{46.76} &  \cellcolor{almond}\textbf{56.47}\\
    % &  & \cellcolor{almond}ALBEF-SS-Ret-g20 (ours) & \cellcolor{almond}26.40 & \cellcolor{almond}47.40 & \cellcolor{almond} 55.00 && \cellcolor{almond}18.37 &\cellcolor{almond} 18.37 &\cellcolor{almond}  18.37\\
 % \cmidrule{2-10}
 %       & \multirow{3}{*}{100} & MedClip &  &  &  &&  38.81 & 54.68 & 54.68\\
 %         &  & MedClip-upto K &  &  &  && 28.47 & 66.11  & 76.55 \\
 %     &  & ALBEF &  &  &  &&  &  &  \\
 %    &  & ALBEF-PT (ours) &  &  &  &&  &  &  \\

    \midrule
    \midrule

 % \multirow{6}{*}{Report-to-Image}    & \multirow{3}{*}{1} & MedCLIP & 0.3 & 26.5  &  47.7 && 1.57 & 26.41 & 62.63 \\
 %  % &  & MedCLIP-PT (ours) &  26.4  & 54.8 & 56.9   &&  &  &  \\
 %  &  & ALBEF &  &  &  &&  &  &  \\
 %    &  & ALBEF-PT (ours) & 24.3 & 59.1 & 71.6 &&  &  &  \\
    
 \multirow{8}{*}{Report-to-Image}    & \multirow{4}{*}{ZS} & MedCLIP-ViT & 24.1 & 42.6 & 46.6 && 35.66 & 55.37 & 81.48\\
 &  & ALBEF-mscoco  & 5.6 & 41.2 & 48.7 && 1.36 & 35.07 & 68.37\\   
  &   & ALBEF-flickr30k & 2.2 & 44.3 & 50.5 && 0.32 & 61.17 & 57.74\\      
 &    & \cellcolor{almond}ALBEF-SS-Ret (ours)& \cellcolor{almond}- &\cellcolor{almond} - &\cellcolor{almond} -  && \cellcolor{almond}\textbf{54.70} & \cellcolor{almond}\textbf{81.94} & \cellcolor{almond}\textbf{85.49}\\
    \cmidrule{2-10}  
   % & \multirow{3}{*}{5} & MedClip & 26.4 & 36.9 & 49.1 & & 36.36 & 59.70   & 81.28 \\
   %  &  & MedClip-upto K &  &  &  && 36.36 & 54.78 & 57.14 \\
   %   &  & ALBEF &  &  &  &&  &  &  \\
   %  &  & ALBEF-PT (ours) &  &  &  &&  &  &  \\
   %  \cmidrule{2-10}

 & \multirow{3}{*}{10} & MedCLIP & 3.3 & 38.6 & 46.4 && 1.57 & 36.64 & \textbf{57.20} \\
   % & \multirow{3}{*}{10} & MedCLIP-old & 26.4 & 32.2 & 41.2 && 31.10 & 39.87 & 63.46 \\
    & &\cellcolor{almond} MedCLIP-SS &\cellcolor{almond} \textbf{6.6} & \cellcolor{almond}33.2 &\cellcolor{almond} \textbf{54.6} &&\cellcolor{almond} \textbf{36.95} & \cellcolor{almond}\textbf{55.53} & \cellcolor{almond}56.68 \\
    % &  &\cellcolor{almond} MedCLIP-SS-g20 (ours) &\cellcolor{almond} 4.8 &\cellcolor{almond} 29.4 &\cellcolor{almond} 59.6 && \cellcolor{almond}36.95 &\cellcolor{almond} 55.22 &\cellcolor{almond} 55.32 \\
    \cmidrule{3-10}
     &  & ALBEF & 32.9 & 65.9 & 75.0 && 36.74 & 68.99 & 81.84 \\
       &  & \cellcolor{almond}ALBEF-SS-Ret & \cellcolor{almond}31.6 & \cellcolor{almond}\textbf{67.3} & \cellcolor{almond}73.2 && \cellcolor{almond}35.39 & \cellcolor{almond}\textbf{78.29} & \cellcolor{almond}80.06\\
    % &  &\cellcolor{almond} ALBEF-SS-Ret-g20 (ours) &\cellcolor{almond} 21.7 &\cellcolor{almond} 63.1 & \cellcolor{almond}72.6 &&\cellcolor{almond} 4.28 &\cellcolor{almond} 42.17 &\cellcolor{almond} 56.99 \\
 % \cmidrule{2-10}
 %       & \multirow{3}{*}{100} & MedClip & 0.1 & 26.9 & 61.6 && 36.36 & 54.68 & 54.68\\
 %           &  & MedClip-upto K &  &  &  && 36,36 & 54.78 & 57.14 \\
 %     &  & ALBEF &  &  &  &&  &  &  \\
 %    &  & ALBEF-PT (ours) &  &  &  &&  &  &  \\

\bottomrule

    \end{tabular}
 \vspace{5pt}
    \caption{\footnotesize Zero-shot (ZS) and few-shot (K=10) results for image$\leftrightarrow$report retrieval. MedCLIP-ViT is pretrained on chest x-rays~\cite{johnson2019mimic}, \cite{irvin2019chexpert}, MedCLIP and MedCLIP-SS are trained on the screening mammogram exams. Numbers are in percentages.}
    \label{tab:zs_fs}
    \vspace{-7mm}
\vspace{-10pt}
\end{table*} 

\noindent \emph{\textbf{Zero shot Image$\leftrightarrow$Text retrieval:} }
We further compare the zero-shot performance on the external test set using off-the-shelf models: MedCLIP-ViT, MSCOCO-pretrained ALBEF, and Flick30K-pretrained ALBEF and compare to ALBEF-SS-Ret pretrained on $\sim$70K internal samples. 
% Off-the-shelf MedCLIP-ViT is pretrained on MIMIC-CXR and CheXpert using 570K samples. 
For image-to-report, MedCLIP-ViT obtains the best R@1 score: 25.7\%  vs. second best 21.61\% from ALBEF-SS-Ret. ALBEF-SS-Ret outperforms MedCLIP-ViT on R@5 and R@10 by 7.61\% $\uparrow$ and 14.43\% $\uparrow$ respectively. For report-to-image retrieval, ALBEF-SS-Ret outperforms MedCLIP-ViT by 19.04\% $\uparrow$, 26.57\% $\uparrow$, and 4.01\% $\uparrow$ in terms of R@1, R@5, and R@10 respectively. See table~\ref{tab:zs_fs} for complete results. 
% We didn't evaluate the zero-shot performance for ALBEF-SS-Ret on the internal dataset since it is trained on the internal data itself. 
% \vspace{-20pt}
%----------------------------------------------------------
\begin{figure}[htb!]
  \centering
\begin{subfigure}{0.46\textwidth}
    \includegraphics[width=\textwidth]{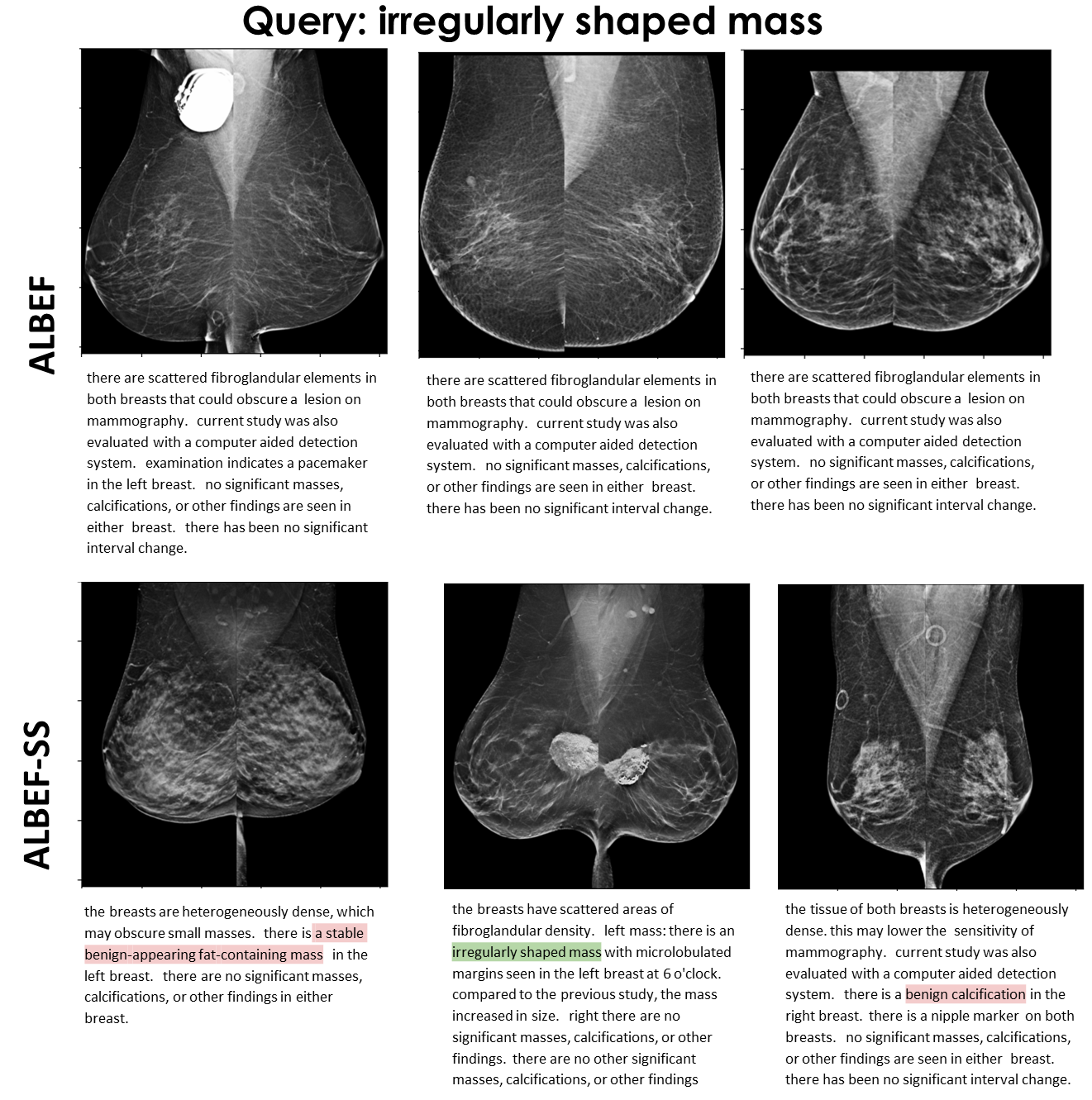} 
    \caption{}
\end{subfigure}
\begin{subfigure}{0.43\textwidth}
    \includegraphics[width=\textwidth]{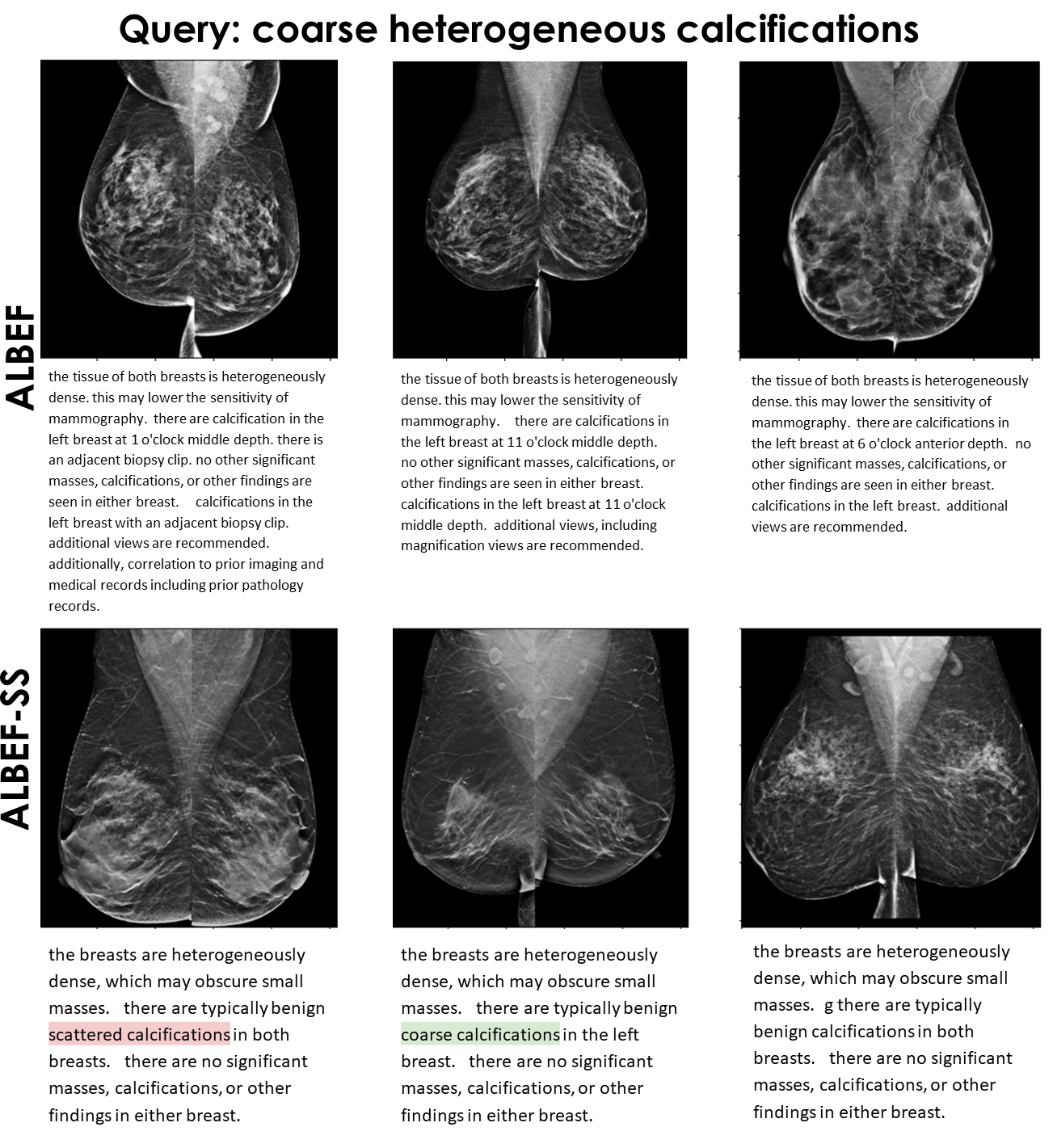} 
    \caption{}
\end{subfigure}
 
  \caption{\footnotesize Qualitative results for Retrieval model. 
  % Query is used to retrieve top-3 relevant cases (left from right) from joint embedding space. 
  An example with highlighted green words is marked relevant by the radiologist for case build. Concepts highlighted with the pink show not exact but related findings in the image-report pair.
  % (a) query for mass and (b) query for coarse calcification. 
  % See the supp. for discussion.
  }
  \label{fig:qual}
  \vspace{-15pt}
\end{figure}
 % \vspace{-5pt}

\noindent \emph{\textbf{Few shot Image$\leftrightarrow$Text retrieval:}}
% To study the effectiveness of selective sampling in a
For the few-shot learning setup, we sampled up to K=10 instances for each group from an internal training set. For groups with less than 10 instances, we keep all available instances. This resulted in 3,331 unique training image-report pairs.
% which is less than 5\% of the original training set.
\textbf{\textit{Internal test set:}} For image-to-report retrieval evaluation, ALBEF-SS-Ret outperforms ALBEF on all three metrics. MedCLIP-SS also demonstrates consistent improvements across all metrics with atleast 50\% relative performance gain over MedCLIP. For report-to-image, MedCLIP shows improvement in R@1 (3.3\%$\uparrow$) and R@10 (8.2\%$\uparrow$). ALBEF-SS-Ret shows overall comparable performance to ALBEF with a slight gain in R@5 score.
% we observe consistent performance gain for R10 using the proposed selective sampling for both MedCLIP-SS and ALBEF-SS. 
\textbf{\textit{External test set:}} 
% When performing external validation, w
We observe that ALBEF-SS-Ret performs significantly better than its counterpart (R@1 score: 20.88\% vs 0.3\%, R@10: 46.76\% vs 29.96\%) when doing image-to-report retrieval during external validation. For report-to-image retrieval, it improves R@5 by approx. 10 points while performing comparable to ALBEF on R@1 and R@10. MedCLIP-SS, in comparison with MedCLIP, also show significant improvement for R@1 (36.95\% vs 1.57\%) and R@5 (55.53\% vs 36.64\%) scores respectively on report-to-image retrieval task, but shows the opposite trend on image-to-report retrieval. Overall, we observe that selective sampling consistently benefits ALBEF model for both internal and external validation. MedCLIP-SS, on the other hand, while being beneficial for internal testing as well as for external validation of report-to-image retrieval performance, seems to be less effective for out-of-domain image-to-report retrieval. This is consistent with the trends observed while performing external validation of MedCLIP-SS when trained on the full training set. 
% For image-to-report retrieval using few-shot, we observed that selective sampling is less effective than usual training. As few-shot training set does not have high class imbalance that existed in the original train set, thus using selective sampling approach does not help. This is intuitive as we are forcing the model to see frequent groups less compared to the usual random sampling. Hence, the performance drop on the test set which still has the large number of samples coming from those groups. As stated before, report-to-image retrieval is comparatively easier task, MedCLIP-SS still observes 5.85\%$\uparrow$ improvement in R1 and 15.35\% $\uparrow$ in R5. The performance however drops for R10 compared to the random sampling. 
We need to re-calibrate the frequent groups to benefit from selective sampling based on the support set's group distribution.
%-----------------------------------------------------------
%  \vspace{-25pt}
% \begin{table}[hbt!]
\begin{table}[t]
\scriptsize
\renewcommand{\arraystretch}{.95}
  \centering \setlength{\tabcolsep}{1.0\tabcolsep}   
    \begin{tabular}{l|ccc|ccc}
       \toprule
      
      &    \multicolumn{3}{c}{Image-to-Report} & \multicolumn{3}{c}{Report-to-Image} \\
    \cmidrule{2-4} \cmidrule{5-7} 
   \multicolumn{1}{c}{\textbf{Method}}  &  R@1 & R@5 & R@10 &  R@1 & R@5 & R@10 \\
      
        \midrule 
    %  (1)  no caps.(X1)     & 62.43 & 11.88 & 65.42 & 20.11 && 1.77 & 9.80 & 3.00 & 34.22 \\
     % (2)  w/o mask       & 95.64 & - & - & - && - & - & - & -\\
     % (1)  B=8 &3.2 & 12.7 & 23.1 & 4.8 & 29.4 & 59.6 \\ 
  
     % (2)  B=32   & 0.3 & 4.6 & 9.1 & 24.4 & 50.3 & 61.5 \\
    
     % (3)  B=48   & 0.2 & 1.6 & 4.0 & 6.6 & 33.1 & 40.2 \\
     %     \midrule 
     (1)  R=0.25  & 0.4 & 1.5 & 2.4 & 3.2 & 29.2 & 41.6 \\
     (2)  R=0.38  & 0.4 & 2.8 & \textbf{8.7}& 15.7 & 30.7 & 51.3 \\
     (3)  R=0.50  & 0.1 & 1.8 & 5.2 & \textbf{17.7} & \textbf{41.2} &  \textbf{58.9}  \\     
     (4) R=0.75  & \textbf{0.5} & 5.2 & 7.7 & 1.4 & 26.3 & 28.9 \\
     \midrule
    %  (14) bert init (C=32, X1)  & - & - & - & - && - & - & - & -\\
    %  (8) pretrained  &  &  &  &|&  &  &  \\
    % (9) fine-tuned  &  &  &  &|&  &  &  \\

      % \midrule
    %  (14) bert init (C=32, X1)  & - & - & - & - && - & - & - & -\\
     (5) w/ B shuffle  & 0.3 & 1.8 & 6.8 & 17.1 & 24.7 & 42.6 \\
    (6) w/o B shuffle & \textbf{0.4} & \textbf{2.8} & \textbf{8.7} & \textbf{15.7} & \textbf{30.7} & \textbf{51.3} \\
 
    % \midrule
    %    (10) MedCLIP, B=8  & 3.2 & 6.0 & 9.9 & 0.4 & 5.5 & 5.5 \\
    % (11) MedCLIP-SS, B=8  & 3.2 & \textbf{12.7} & \textbf{23.1} & \textbf{4.8} & \textbf{29.4} & \textbf{59.6} \\
    % \hline
       \midrule
       (7) Freq. groups, fixed & 17.00 & 44.30 & 55.30 & 32.90 & 66.50 & 73.80 \\
    (8) Freq. groups, recalibrate  & \textbf{25.40} & \textbf{48.10} & \textbf{57.40} & 31.60 & \textbf{67.30} & 73.20 \\
 
	\bottomrule
    \end{tabular}
    % \vspace{-5pt}
    \caption{\footnotesize Ablations for the proposed sampling strategy on Institute X using MedCLIP-SS model. B=batch size, R=ratio of frequent groups to rare groups in a batch, 
    % PT=pretrained model to extract support set features, FT=fine-tuned using support set, K shots for few shot learning.
    % All models were trained using few shot learning with K=10 except row (10) and (11). 
    % Results for the final design choices are shown in bold. 
    % See section~\ref{sec:ablations} for discussion.
    % Numbers are in percentages.
    }
    \vspace{-15pt}
    \label{tab:ablations_mean}
    \vspace{-0.5cm}
\end{table} 
%----------------------------------------------------------------------
% \begin{figure*}[t]
% \begin{tabular}{cccc}
% % \subcaptionbox{ITA Loss curves}{\includegraphics[width=0.4\linewidth]{figures/before_vs_after_ita_loss.pdf}} &
% % &

% \subcaptionbox{Internal test set}{\includegraphics[width=0.45\linewidth]{figures/uwtest_groups_piechart.png}} &
% \subcaptionbox{Externel test set}{\includegraphics[width=0.45\linewidth]{figures/mayotest_groups_piechart.png}} 
% \end{tabular}

%  \caption{\footnotesize Groups distribution for internal and external test sets.
%  % For brevity, rare groups are grouped as `rare groups'. 
%  For both test sets, top 3 groups belong to breast composition. Breast tissue composition could be scattered fibroglandular (S), heterogeneous (H), fatty (F), and extreme dense (E). Short forms are used for asymmetry (asymm) and calcifications (calc). }
% \label{fig:test_data_dist}
% \vspace{-5pt}
% \end{figure*}
\paragraph{Ablations and Analyses:} \label{sec:ablations}
% We designed the ablations to understand effect of batch formation strategy and distribution of frequent and rare groups upon the proposed selective sampling.
We used MedCLIP-SS with few-shot learning (K=10) in all ablations unless specified otherwise. Table~\ref{tab:ablations_mean} reports the selected ablations from our detailed analyses regarding important hyperparameters such as \#samples from frequent vs. rare groups, recalibrating no. of frequent groups with change in data distribution that happens during few-shot learning, and choice of mini-batch shuffling after our selective sampling. See supp. document for details.
\vspace{-10pt}
\section{Discussion and Conclusion}
\vspace{-5pt}
% In this work, we present a knowledge grounded selective sampling strategy to train vision language models (VLMs) on medical data with extreme class imbalance and presence of rare anomaly samples.
Training a large network on medical data, particularly with contrastive loss, is always challenging when the dataset is highly influenced by the majority `normal' cases and instances with compelling representation (image or textual) are extremely rare. Moreover, contrastive loss can be affected by the quality and diversity of the negative pairs, which can be hard to sample from a large and complex dataset. Our proposed knowledge-grounded selective sampling strategy helps the contrastive model training by ensuring the sampling of the true negatives and equalize representation of rare cases.  
% We propose a mechanism to extract knowledge from textual radiology reports of screening mammogram, perform knowledge-grounded grouping on image-report pairs using the extracted knowledge, and exploit the resulting grouped data to train VLMs. 
% We demonstrate the effectiveness of the proposed selective sampling by training two vision language models ALBEF and MedCLIP and evaluating across both internal and external datasets. 
We observed improvement in the retrieval performance with the selective sampling strategy, especially for the ALBEF model. For MedCLIP, we observed improvement for internal evaluation; however there was no improvement on the external dataset for image-to-report which could be based on the fact that image-to-text retrieval is a more challenging task and we didn't pre-train the MedCLIP on the mammogram dataset. However, we still observed MedCLIP performance improvement on the external dataset for report to image particularly in R@1 and R@5 for few-shot learning.
On the zero-shot performance, our pre-trained model also outperformed all the baselines, including MedCLIP-VIT,  on the external dataset for both image-to-report and report-to-image retrieval task. 
% While zero-shot performance clearly shows the gain of pre-training and selective sampling, there is an inconsistent performance gain for few-shot learning in image-to-report retrieval which could be due to the fact that extreme class imbalance is not impactful for few-shot.
% as we are already presenting classic examples of rare cases during the few-shot sampling. 
It is also highlighted in the domain of LLMs that  few-shot learning can be highly sensitive to the quality of the demonstrations, emphasizing the need
for strategies to strategically select few-shot~\cite{zhao2021calibrate}.
% Theoretically, to improve the performance also on the few-shot we have to balance the number of knowledge group representation but this is not within the scope of this work. 

Based on the ablation study, we also present the fact that proposed selective sampling can help to train the VLM model with smaller batch size for a limited resource setting. However, thorough experimentation needs to be done with intelligent sampling to balance the groups for larger batch size to properly understand the relationship between the number of groups and the batch size. 
% However, due to the time and resource constraints, we are unable to conclude the experiment.    

In summary, our proposed sampling strategy lays the groundwork to rethink data sampling strategies for effective training of multimodal networks as well as for in-context learning, case in point, vision-language models grounded in the multimodal data for medical contexts.

%
% ---- Bibliography ----
%
% BibTeX users should specify bibliography style 'splncs04'.
% References will then be sorted and formatted in the correct style.
%
\bibliographystyle{splncs04}
\bibliography{references}

\appendix

% This is samplepaper.tex, a sample chapter demonstrating the
% LLNCS macro package for Springer Computer Science proceedings;
% Version 2.20 of 2017/10/04
%
% \documentclass[runningheads]{llncs}
% %
% \usepackage{multirow} 
% % Include other packages here, before hyperref.
% \usepackage{graphicx}
% \usepackage{booktabs}
% \usepackage{xcolor, colortbl}
\definecolor{almond}{RGB}{186,210,225}

% \usepackage[table,xcdraw]{xcolor}
% \usepackage{subcaption}
% Used for displaying a sample figure. If possible, figure files should
% be included in EPS format.
%
% If you use the hyperref package, please uncomment the following line
% to display URLs in blue roman font according to Springer's eBook style:
% \renewcommand\UrlFont{\color{blue}\rmfamily}

% \begin{document}
%
\title{Supplementary: Knowledge-grounded Adaptation Strategy for Vision-language Models: Building Unique Case-set for Screening Mammograms for Residents Training}
%
%\titlerunning{Abbreviated paper title}
% If the paper title is too long for the running head, you can set
% an abbreviated paper title here
%

\author{}
%
% \authorrunning{Paper ID 3702}
% First names are abbreviated in the running head.
% If there are more than two authors, 'et al.' is used.
%
\institute{
% Department of Radiology, Mayo Clinic \and
% Department of Radiology, UW Madison School of Medicine and Public Health \and
% School of Computing and Augmented Intelligence, Arizona State University
}

% \author{Aisha Urooj Khan\inst{1} \and
% John Garrett \inst{2} \and
% Tyler Bradshaw\inst{2} \and
% Lonie Salkowski\inst{2} \and
% Jiwoong Jason Jeong\inst{3} \and
% Amara Tariq \inst{1} \and
% Imon Banerjee\inst{1,3}}
% %
% \authorrunning{A. Urooj et al.}
% % First names are abbreviated in the running head.
% % If there are more than two authors, 'et al.' is used.
% %
% \institute{
% Department of Radiology, Mayo Clinic \and
% Department of Radiology, UW Madison School of Medicine and Public Health \and
% School of Computing and Augmented Intelligence, Arizona State University
% }

% }
%
\maketitle              % typeset the header of the contribution

In this document, we discuss related work, ablations and qualitative analyses, and additional discussion about the proposed approach.

\section{Related Work}
\emph{i. Vision-language model in radiology} - Automated medical report generation from radiology images are one of most popular task for VLM. Nooralahzadeh et al.~\cite{nooralahzadeh2021progressive} proposed a two-step model which derived global concepts from the image then reformed them into finer and coherent texts using a transformer architecture. You et al.~\cite{you2021aligntransformer} proposed AlignTransformer where they implemented align hierarchical attention (AHA) and multi-grained transformer (MGT) to produce the disease tag for templated report generation without considering uncertainty of the findings. Wang et al.~\cite{wang2024trust} proposed a confidenece guided method for VLM which explicitly quantified visual and textual uncertainties for radiology report generation. Alfarghaly et al.~\cite{alfarghaly2021automated} proposed a deep learning model consisting of CNN-based Chexnet model as encoder and a Transformer model as decoder. Similar to You et. al.~\cite{you2021aligntransformer}, they used Chexnet to predict the tags for images and also to generate latent space vector. Finally to generate a report, they used a GPT2 pre trained model on the latent space vetor and semantic features. Mohsan et. al.~\cite{mohsan2022vision} used a pre-trained vanilla image transformer architecture and combine it with different pre-trained language transformers as decoder to generate chest X-ray reports. Most of current VLM models in radiology are focused on 2D chest X-ray. Given the scarcity of the open-source mutlimodal dataset (reports+images) and the complexity of processing mammogram images(large dimension, varying density, mutli-view), VLM literature is limited in mammogram domain.

\noindent\emph{ii. Multi-modal Retrieval in radiology} - Multimodal retrieval framework can help in the case-building with simple text description or similar image search. Content-based image retrieval and simple `key-word' based text retrieval are the most widely used retrieval mechanism in radiology. Recently, multimodal retrieval using image-text contrastive pre-training is gaining interest. CXR-RePaiR\cite{endo2021retrieval} adopts a contrastive image-text retrieval method that retrieves a report whose CLIP~\cite{radford2021learning} text embedding scores the highest cosine similarity with the chest X-ray’s CLIP image embedding where CLIP uses contrastive imge-language pre-training. Jeong et. al.~\cite{xrem} proposed X-REM that uses an image-text matching score using a multimodal encoder to measure the similarity of a chest X-ray image and radiology report for report retrieval. ConVIRT~\cite{zhang2022contrastive} jointly trains the vision and text encoders with the paired medical images and reports via a bidirectional contrastive objective; GLoRIA~\cite{huang2021gloria} further models both the global and local interactions between medical images and reports to capture the pathology meanings from specific image region. MedClip~\cite{medclip} replaced InfoNCE loss with semantic matching loss based on medical knowledge~\cite{lazarus2006bi} to eliminate false negatives in contrastive training of the VLMs. However, current multi-modal retrieval frameworks have significant limitations - (i) no strategy proposed to preserve `rare' case representation which is extremely important in generating meaningful embedding space for minority samples; (ii) mining `hard-negatives' is challenging in radiology, particularly for mammogram case-studies giving most templated reports are syntactically similar with limited concept difference while images presents distinct features not mentioned/partially mentioned in the reports.

% \subsection{VLM Training}
% \vspace{-5pt}
% The above mentioned sampling strategy can be used to sample mini-batches to train the vision-language model for image-text alignment. We assess the effectiveness of the sampling strategy in two settings: pretraining and few-shot learning across two existing VLMs: ALBEF~\cite{ALBEF} and MedCLIP~\cite{medclip}.
% ALBEF and MedCLIP are two-stream VLMs whereas ViLT has a single encoder for images and text. 
\vspace{-5pt}
\subsection{Evaluation Metrics}
\vspace{-5pt}
On the joint embedding space, we measure the retrieval performance separately on both text and image query. To measure the performance, we consider Recall@K metric assessing top-1, top-5, and top-10 performance. Because of the groups we construct by preprocessing report findings, we consider a hit if any report with the same findings (hence the same group) appears in the top-K ranks. For few-shot experiments, we consider K=10 shots to finetune the models.

\subsection{More details about datasets:}
The internal dataset Institute Y's population of 58.7 average age (median age: 59, interquartile range: 15 [51, 66]) includes ~92\% white, ~3\% black, 2.5\% Asian, and the remaining 2.5\% are other/unknown. 
The exams considered for this work contains digital breast tomosynthesis (DBT) combined with digital mammography, and we selected Left-MLO and Right-MLO 2D view from the digital mammography. See figure 1 (supp) for group distributions of internal and external test sets. The distribution for both institutes is not very different despite of template-based radiology reports for institute X, and free-form text reports for institute Y.

%----------------------------------------------------------------------

\subsection{Additional Implementation Details}

\paragraph{\textbf{ALBEF}~\cite{ALBEF}:} ALBEF has an image encoder and a text encoder, followed by a cross encoder with image-text contrastive loss. An image-text alignment loss is used to align image and text features even before cross-attention. Image-Text Matching (ITM) and Masked LM (MLM) are used in addition to jointly optimize the model. The baseline ALBEF model initializes from DeiT~\cite{deit} vision transformer using $16\times16$ patch size. We pretrain ALBEF on Institute X image-report pairs, followed by a retrieval-only task ITM for fine-tuning the pretrained backbone. For a $512\times512$ image and the patch size of $16\times16$, vision encoder takes $1024$ patch tokens 
% ($512\div16=32;32\times32=1024$)
in the ALBEF model. We train ALBEF with (ALBEF-SS) and without (ALBEF-Ret) the proposed selective sampling. We validated the model on the Institute Y external dataset. 

\paragraph{\textbf{MedCLIP}~\cite{medclip}:} MedCLIP is a variant of CLIP~\cite{radford2021learning}  model is originally trained on gray-scale chest x-rays and reports. We evaluate MedCLIP pretrained on CheXPERT dataset~\cite{irvin2019chexpert}  and MIMIC-CXR~\cite{johnson2019mimic} for zero-shot, initialize model weights for few shot learning, and train MedCLIP from scratch on the 2D mammogram images for fully supervised backbone. Similar to ALBEF, we also trained MedCLIP with (MedCLIP-SS) and  
without (MedCLIP) the proposed selective sampling. We used the same internal and external validation sets.
For full training, we consider top 20 groups w.r.t number of samples as frequent groups out of total 1005 unique groups. Ratio of frequent to rare groups in a minibatch is set to R=0.375. We use batch size=8 and boundary b=3 for random sampling of frequent and rare groups, i.e., for R=0.375 - 5 samples belong to frequent groups, and 3 are sampled from the set of rare groups. All training parameters except the hyperparameters considered for this study stay the same across models. We finetuned the MedCLIP on internal training set using their multiclass task. However, we observe that multiclass task learning tends to hurt image-to-report performance. 

% \subsubsection{ViLT~\cite{vilt}:} ViLT is a single multimodal encoder based model, a ViT~\cite{} like model trained on concatenated image patch tokens and paired text tokens. We fine-tune the model using their vanilla approach and compare with ViLT-SS trained with the proposed selective sampling. 

%----------------------------------------------------------------------
\begin{figure*}[t]
\begin{tabular}{cccc}
% \subcaptionbox{ITA Loss curves}{\includegraphics[width=0.4\linewidth]{figures/before_vs_after_ita_loss.pdf}} &
% &

\subcaptionbox{Internal test set}{\includegraphics[width=0.45\linewidth]{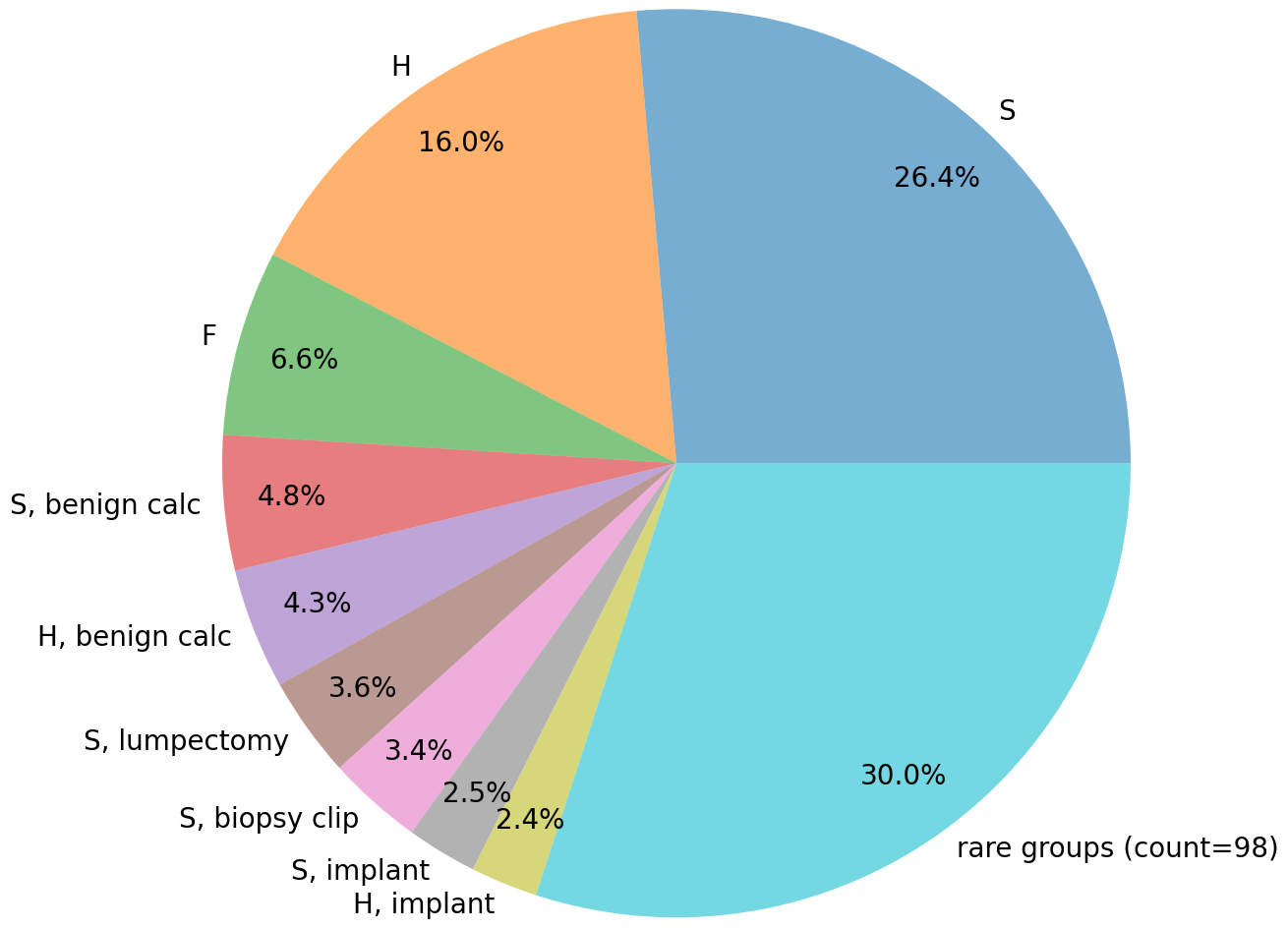}} &
\subcaptionbox{Externel test set}{\includegraphics[width=0.45\linewidth]{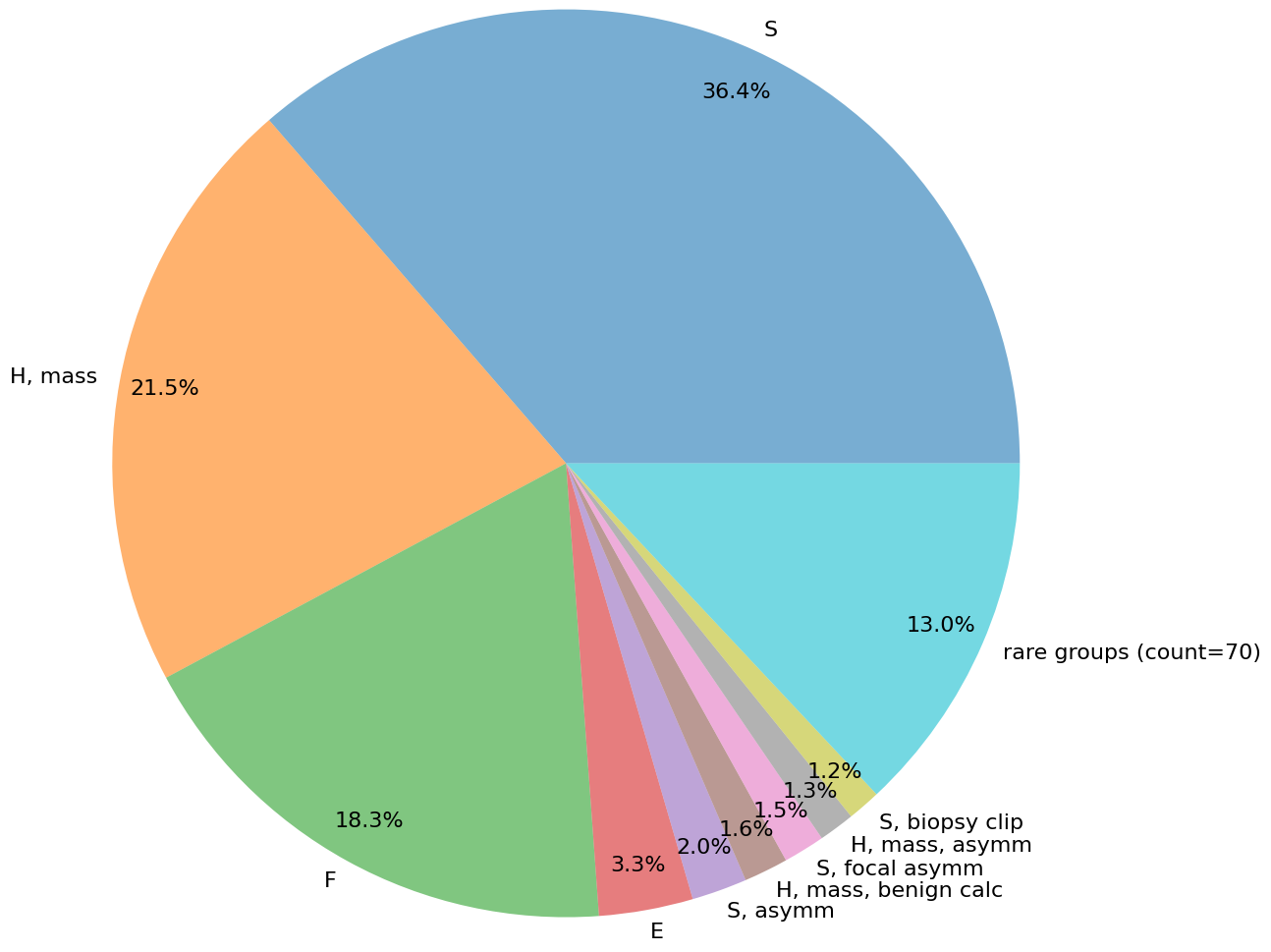}} 
\end{tabular}

 \caption{\footnotesize Groups distribution for internal (institute X) and external (institute Y) test sets.
 % For brevity, rare groups are grouped as `rare groups'. 
 For both test sets, top 3 groups belong to breast composition. Breast tissue composition could be scattered fibroglandular (S), heterogeneous (H), fatty (F), and extreme dense (E). Short forms are used for asymmetry (asymm) and calcifications (calc). }
\label{fig:test_data_dist}
\vspace{-5pt}
\end{figure*}

\begin{figure*}[htb!]
\centering
% \begin{tabular}{cccc}
% \subcaptionbox{ITA Loss curves}{\includegraphics[width=0.4\linewidth]{figures/before_vs_after_ita_loss.pdf}} &
% &
\includegraphics[width=0.8\linewidth]{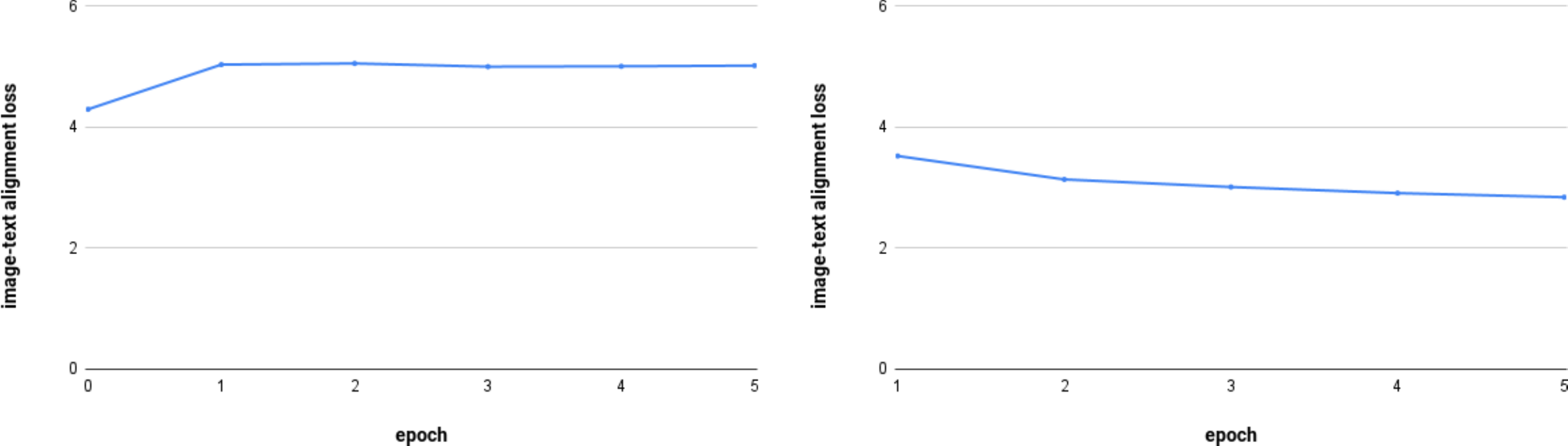}
% \subcaptionbox{}{\includegraphics[width=0.45\linewidth]{figures/uwtest_groups_piechart.png}} &
% \subcaptionbox{}{\includegraphics[width=0.45\linewidth]{figures/mayotest_groups_piechart.png}} 
% \end{tabular}

 \caption{\small Loss curves for image-text alignment loss in ALBEF~\cite{ALBEF}. Left) vanilla ALBEF trained on internal dataset, Right) ALBEF after using proposed selective sampling. }
\label{fig:before-vs-after}
\vspace{-5pt}
\end{figure*}
\vspace{-15pt}

%-----------------------------------------------------------
%  \vspace{-25pt}
% \begin{table}[hbt!]
\begin{table}[t]
\small
\renewcommand{\arraystretch}{.95}
  \centering \setlength{\tabcolsep}{1.0\tabcolsep}   
    \begin{tabular}{l|ccc|ccc}
       \toprule
      
      &    \multicolumn{3}{c}{Image-to-Report} & \multicolumn{3}{c}{Report-to-Image} \\
    \cmidrule{2-4} \cmidrule{5-7} 
   \multicolumn{1}{c}{\textbf{Method}}  &  R@1 & R@5 & R@10 &  R@1 & R@5 & R@10 \\
      
        \midrule 
    %  (1)  no caps.(X1)     & 62.43 & 11.88 & 65.42 & 20.11 && 1.77 & 9.80 & 3.00 & 34.22 \\
     % (2)  w/o mask       & 95.64 & - & - & - && - & - & - & -\\
     (1)  B=8 &3.2 & 12.7 & 23.1 & 4.8 & 29.4 & 59.6 \\ 
      
    %  (3)  e2e(C=32,X1)    & 57.02 & 10.36 & 64.28 & 17.84 && 1.33 & 8.30 & 2.29 & -\\
    %  (4)  2-stage(C=32,X1) & 55.41    & 10.09 & 71.95 & 17.70 && 1.41 & 10.09 & 2.47 & -\\
     %(5)  2-stage, freeze wts(X1)  & - & - & - & - && - & - & -& -\\
     (2)  B=32   & 0.3 & 4.6 & 9.1 & 24.4 & 50.3 & 61.5 \\
    
     (3)  B=48   & 0.2 & 1.6 & 4.0 & 6.6 & 33.1 & 40.2 \\
         \midrule 
     (4)  R=0.25  & 0.4 & 1.5 & 2.4 & 3.2 & 29.2 & 41.6 \\
     (5)  R=0.38  & 0.4 & 2.8 & \textbf{8.7}& 15.7 & 30.7 & 51.3 \\
     (6)  R=0.50  & 0.1 & 1.8 & 5.2 & \textbf{17.7} & \textbf{41.2} &  \textbf{58.9}  \\     
     (7) R=0.75  & \textbf{0.5} & 5.2 & 7.7 & 1.4 & 26.3 & 28.9 \\
     \midrule
    %  (14) bert init (C=32, X1)  & - & - & - & - && - & - & - & -\\
    %  (8) pretrained  &  &  &  &|&  &  &  \\
    % (9) fine-tuned  &  &  &  &|&  &  &  \\

      % \midrule
    %  (14) bert init (C=32, X1)  & - & - & - & - && - & - & - & -\\
     (8) w/ B shuffle  & 0.3 & 1.8 & 6.8 & 17.1 & 24.7 & 42.6 \\
    (9) w/o B shuffle & \textbf{0.4} & \textbf{2.8} & \textbf{8.7} & \textbf{15.7} & \textbf{30.7} & \textbf{51.3} \\
 
    \midrule
       (10) MedCLIP, B=8  & 3.2 & 6.0 & 9.9 & 0.4 & 5.5 & 5.5 \\
    (11) MedCLIP-SS, B=8  & 3.2 & \textbf{12.7} & \textbf{23.1} & \textbf{4.8} & \textbf{29.4} & \textbf{59.6} \\
    % \hline
       \midrule
       (12) Freq. groups, fixed & 17.00 & 44.30 & 55.30 & 32.90 & 66.50 & 73.80 \\
    (13) Freq. groups, recalibrate  & \textbf{25.40} & \textbf{48.10} & \textbf{57.40} & 31.60 & \textbf{67.30} & 73.20 \\
 
	\bottomrule
    \end{tabular}
    \vspace{5pt}
    \caption{\footnotesize Ablations over the design choices for the proposed sampling strategy on Institute X using MedCLIP-SS model. B=batch size, R=ratio of frequent groups to rare groups in a batch, 
    % PT=pretrained model to extract support set features, FT=fine-tuned using support set, K shots for few shot learning.
    All models were trained using few shot learning with K=10 except row (10) and (11). 
    Results for the final design choices are shown in bold. See section~\ref{sec:ablations} for discussion. Numbers are in percentages.}
    \vspace{-15pt}
    \label{tab:ablations_mean}
    % \vspace{-0.5cm}
\end{table}

\subsubsection{Ablations and Analyses:} \label{sec:ablations}

Here, we discuss the ablation results presented in the main paper along with some additional experiments. For convenience, we show the ablation table again in the supplementary. We designed the ablations to understand effect of batch formation strategy and distribution of frequent and rare groups upon the proposed selective sampling. We used MedCLIP-SS with few-shot learning (K=10) in all ablations unless specified otherwise. 
% \paragraph{\textbf{Pretrained model vs. finetuning for few shot learning:}} 
% % discuss why mean features from a pretrained model can make the task harder
% We explored two ways for few-shot learning: 1) using pretrained backbone to obtain features for support set; and 2) use support set to fine-tune the model. For pretrained model, features from each group are averaged to get a mean feature representing each group. For cross-modal retreival, the similarity is computed between the test query features and support set features to find the closest match. If the correct group from support set appears in the top-K, its considered a hit. Keeping only mean features from the support set makes the retrieval task harder compared to group-based retrieval within test set, hence, we notice the drop in performance achieving R@1:[\%], R@5: [\%], and R@10: [\%]. Instead, we use the simpler approach to evaluate few-shot fine-tuned model on our test set. This also allows the evaluation criteria consistent among all methods.
\vspace{-5pt}
\paragraph{\textbf{Impact of batch size:}}
With selective sampling (SS), the model's performance is better for smaller mini-batches. We trained the model for batch size of 8, 32, 48, and 64, and find that B=8 yields best results with SS. However, with increase in the batch size, we also need to adjust the boundary to include more rare groups. We discuss the impact of boundary $b$ in table~\ref{freq_vs_rare}. To study the impact of batch size, we kept the ratio of frequent groups and rare groups same, i.e., R=0.375. In table~\ref{tab:ablations_mean}, row 10 and 11 show results with B=8 for full training of MedCLIP vs. MedCLIP with selective sampling. We observe that using small batch size severely hurts MedCLIP's performance while small batch size helps in MedCLIP-SS. This also highlights the fact that using SS we can train the VLM with smaller batch size in a limited resource setting. 
% direct batch size, relation of batch size and boundary b
\vspace{-5pt}
\paragraph{\textbf{No. of samples from frequent vs. rare groups:}} \label{freq_vs_rare}
Boundary $b$ decides how many samples should come from the rare groups. To study the impact of boundary b (hence the variation in ratio R), we train the MedCLIP-SS model of batch size B=32 with different boundaries. The boundary is determined with ratio R as follows: $b = \lceil B \times R \rceil$. For B=32, we trained with $R \in \{0.25, 0.375, 0.5, 0.75\}$. We find that R=0.375 and R=0.5 gives us better results without any clear winner. We use R=0.375 for results in our main table.
\vspace{-5pt}
\paragraph{\textbf{Recalibrating no. of frequent groups:}} \label{num_freq_groups}
With the change in class distribution of imbalanced data, selection of frequent groups also require modification. For the original group distribution, top 20 groups were selected as frequent based on the knowledge that they cover $\sim$80\% of the train set. For few-shot learning, as we select upto K samples per group (class), all groups with at least K samples are treated equally. Hence, we need to adjust the number of groups used as frequent groups. We trained ALBEF-SS in two settings: 1) keeping the same numger of frequent groups as full training, and 2) adjusting the separation boundary between frequent groups and rare groups. For few-shot support set with K=10, any group with less than 5 samples is considered as a rare group. This results in 222 unique groups as frequent classes, and 783 rare groups out of the total 1005 groups in training data. Readjusting the number of frequent groups helped in image-to-group retrieval task. This recalibration improved R1, R5, and R10 over the baseline (table~\ref{tab:ablations_mean}, row 12) by 8\%, 3.8\%, and 2.1\% respectively while achieving comparable performance for report-to-image retrieval. There is a possibility that further readjustment of number of frequent groups may have improved the performance even better. But this experiment provides us the proof of concept that readjustment in number of frequent groups and rare groups will be needed based on the class distribution in a given dataset to get benefit of selective sampling. 

% --need to discuss impact of hyperparameter boundary b, this boundary b is related to the ratio thing
% --maybe combine these two in one discussion later 

% \paragraph{\textbf{Ratio of frequent:rare groups in a mini-batch:}}
% --impact of ratio of frequent:rare groups in a batch, i.e., 37\%, 50\%, 75\%
% one might argue that frequent groups should have a higher fraction because \#rare groups is much higher than frequent groups -- case in point: 20 most frequent groups in uwmadison covers 82\% of data.
\vspace{-5pt}
\paragraph{\textbf{Mini-batch shuffling after selective sampling:}}
We randomly sample image-report pairs from frequent and rare groups with a fixed boundary $b$, i.e, first B-b samples come from frequent groups, and b samples from rare groups. To study whether shuffling after sampling is helpful or not, we train a model with shuffling again after batch sampling.
% to get a mixture of sampled frequent and rare groups. 
Surprisingly, we find that keeping that shuffling yields lower performance compared to keeping the boundary fixed. For Image-to-Report, we obtain R10=6.8\% vs. 8.7\% ($\sim 2\% \uparrow$) for shuffling after sampling vs. not shuffling. In Report-to-image retrieval, we obtain R10=42.6\% vs. 51.3\% ($8.7\% \uparrow $) respectively.
\begin{table}[t]
\small
\renewcommand{\arraystretch}{.95}
  \centering \setlength{\tabcolsep}{1.0\tabcolsep}   
    \begin{tabular}{ll}
       \toprule
      
  Groups    &  Frequency \\

        \midrule 
   scattered fibroglandular densities &264 \\
 heterogeneously dense & 160\\
 fatty& 66\\
 scattered fibroglandular densities, benign calcification& 48\\
 benign calcification, heterogeneously dense &43\\
 scattered fibroglandular densities, lumpectomy& 36\\
 biopsy clip, scattered fibroglandular densities& 34\\
 scattered fibroglandular densities, implant& 25\\
 implant, heterogeneously dense &24\\
 biopsy clip, heterogeneously dense& 23\\
 fatty, benign calcification &20\\
 lumpectomy, heterogeneously dense &17\\
 scattered fibroglandular densities, asymmetry &11\\
 biopsy clip, scattered fibroglandular densities, benign calcification& 10\\
 scattered fibroglandular densities, focal asymmetry& 10\\
 extremely dense& 10\\
 focal asymmetry, heterogeneously dense& 9\\
 mass, heterogeneously dense& 9\\
 benign calcification vascular, scattered fibroglandular densities& 8\\
 reduction, scattered fibroglandular densities& 8\\
	\bottomrule
    \end{tabular}
    \vspace{5pt}
    \caption{\footnotesize Top 20 groups in the internal test set.}
    \vspace{-15pt}
    \label{tab:ablations_mean}
    % \vspace{-0.5cm}
\end{table}

\paragraph{\textbf{Training loss curves before and after selective sampling:}} In figure 2 (supp.), we show the training loss curves for the ALBEF model before and after selective sampling. We can see that without selective sampling, the image-text alignment loss was actually increasing. Our proposed selective sampling resolves that problem and largely improves the joint embeddings as shown in the results.

% currently boundary was stayed fixed in experiments, does it have any impact on training, i.e., does model gets some kind of bias based on the samples position?
%----------------------------------------------------------
\begin{figure}[tb!]
  \centering
\begin{subfigure}{0.46\textwidth}
    \includegraphics[width=\textwidth]{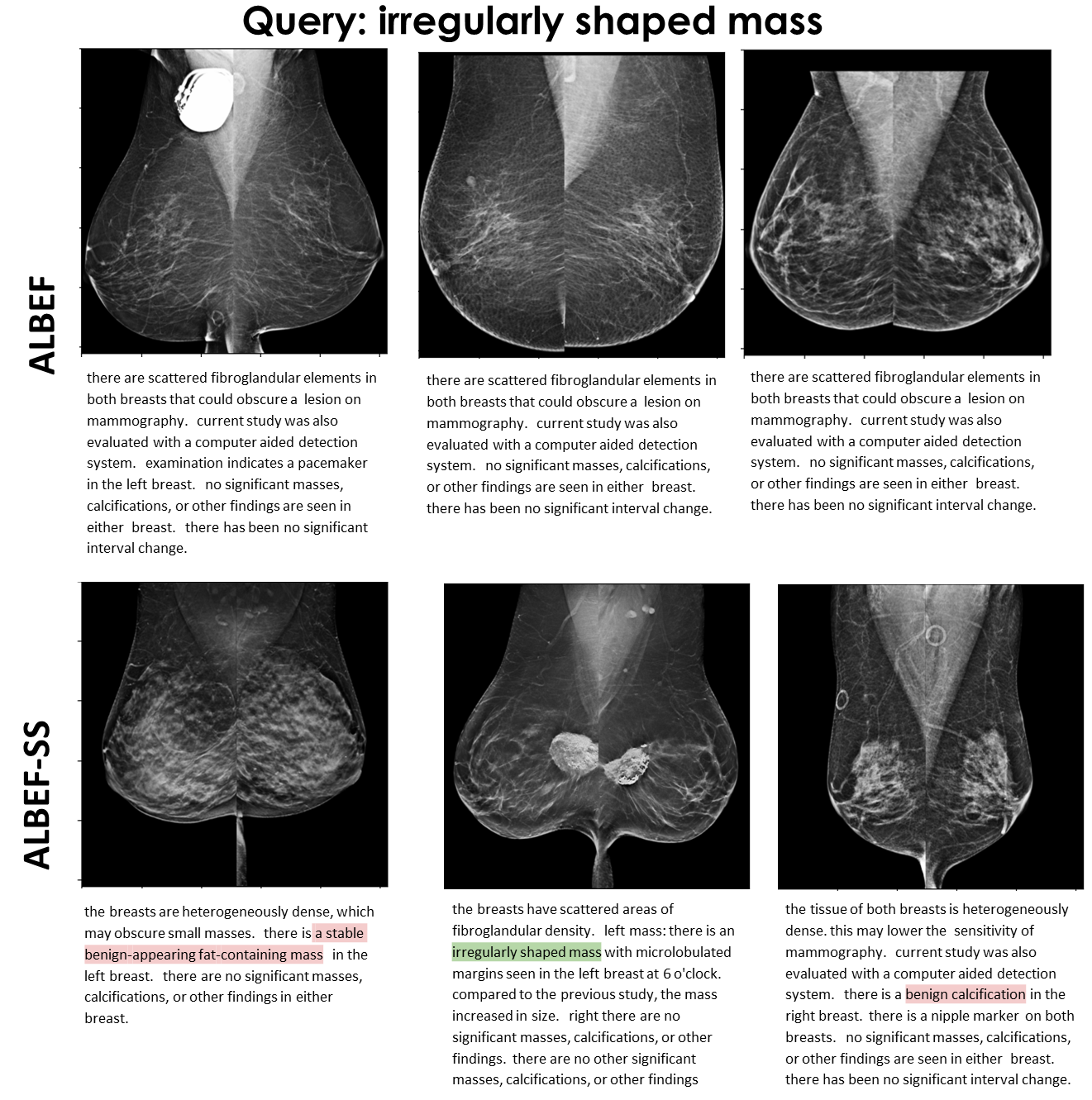} 
    \caption{}
\end{subfigure}
\begin{subfigure}{0.43\textwidth}
    \includegraphics[width=\textwidth]{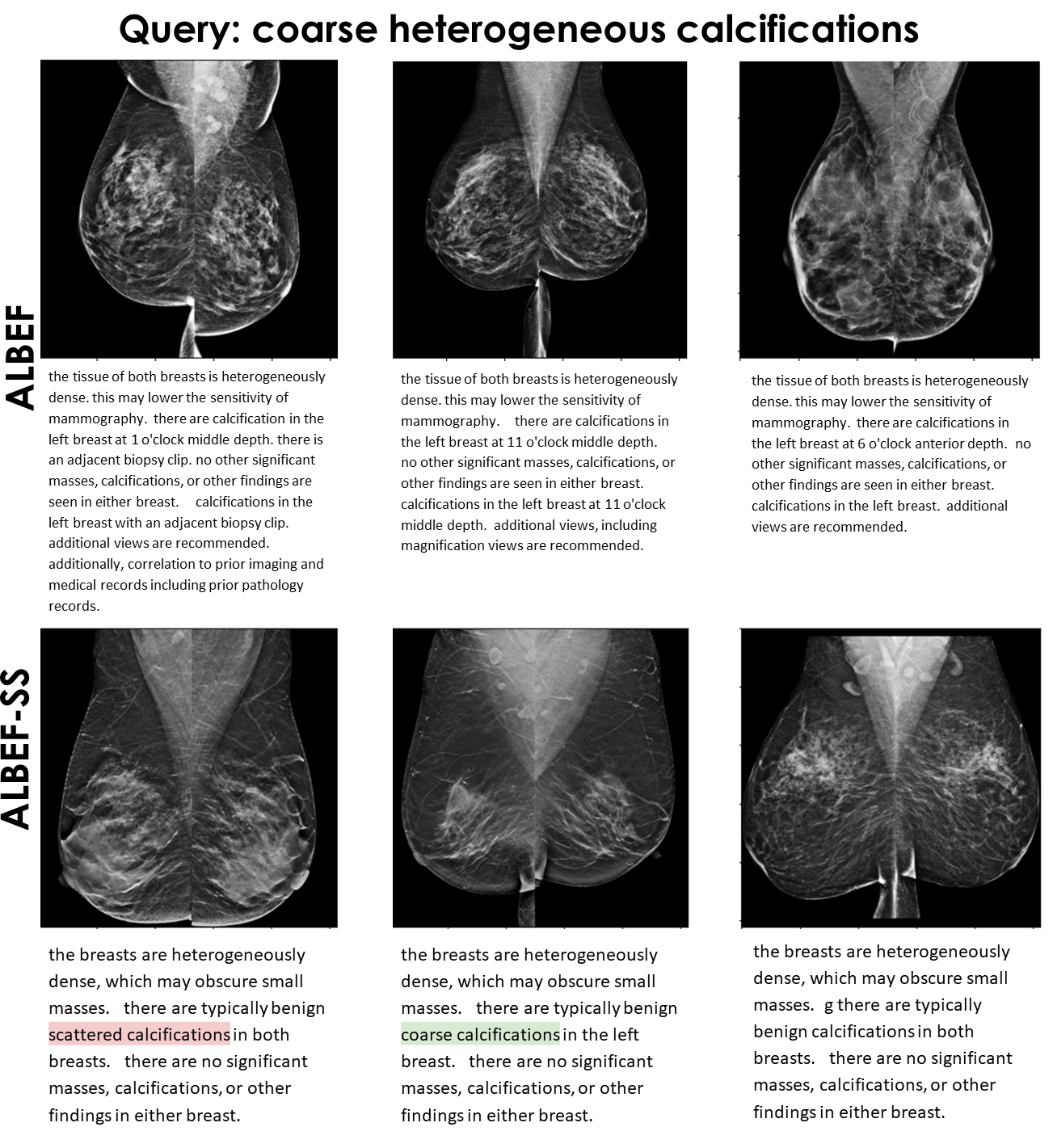} 
    \caption{}
\end{subfigure}
 
  \caption{Qualitative results for Retrieval model. Query is used to retrieve top-3 relevant cases (left from right) from joint embedding space. Example with highlighted green words is marked relevant by radiologist for case build. Concepts highlighted with pink show the not exact but related finding in the image-report pair. (a) query for mass and (b) query for coarse calcification. See the discussion in this document.}
  \label{fig:qual}
  \vspace{-5pt}
\end{figure}
 \vspace{-5pt}

\subsubsection{Qualitative Analysis:} Figure 1 in the main paper shows a session with radiologist and in figure 3 (also shown here), we show results of query-based retrieval on joint-embedding for simulation case build. Top-3 results are shown from left to right. For query `\textit{irregularly shaped mass}', ALBEF without selective sampling retrieves the `no finding' case with the same tissue density, `scattered fibroglandular density'.
% a most common case in the training data. However, all retrieved examples have the same tissue density, i.e., `scattered fibroglandular density'. 
The breast composition, however, is an easy concept to learn from mammograms, i.e.,
% It indicates the retrieved results have the similar semantic representation based on tissue density that is an easy concept to learn from mammograms. 
 figure~\ref{fig:test_data_dist} shows the test set groups' distribution where top-3 groups belong to breast composition. Using selective sampling, the relevant result as marked by a radiologist is fetched in top-3 cases. 
 % We also notice that each of the retrieved results have identified mass or calcification in it.
 The top-1 image-report pair shows `a stable benign-appearing mass', however, the best matched result according to a trained breast radiologist's evaluation is the second case. This shows the challenging nature of this fine-grained retrieval task for screening mammogram. In the second query `\textit{coarse heterogenous calcifications}', the baseline model was able to understand the concept of calcifications (row 1, columns 4-6), but doesn't retrieve results based on the calcification's sub-type, i.e., coarse calcification. ALBEF-SS is able to retrieve the correct image-report pair with `coarse calcifications' (highlighted in green, row 2, column 5).

\begin{figure*}[t]
\begin{tabular}{cccc}
% \subcaptionbox{ITA Loss curves}{\includegraphics[width=0.4\linewidth]{figures/before_vs_after_ita_loss.pdf}} &
% &

\subcaptionbox{ALBEF}{\includegraphics[width=0.45\linewidth]{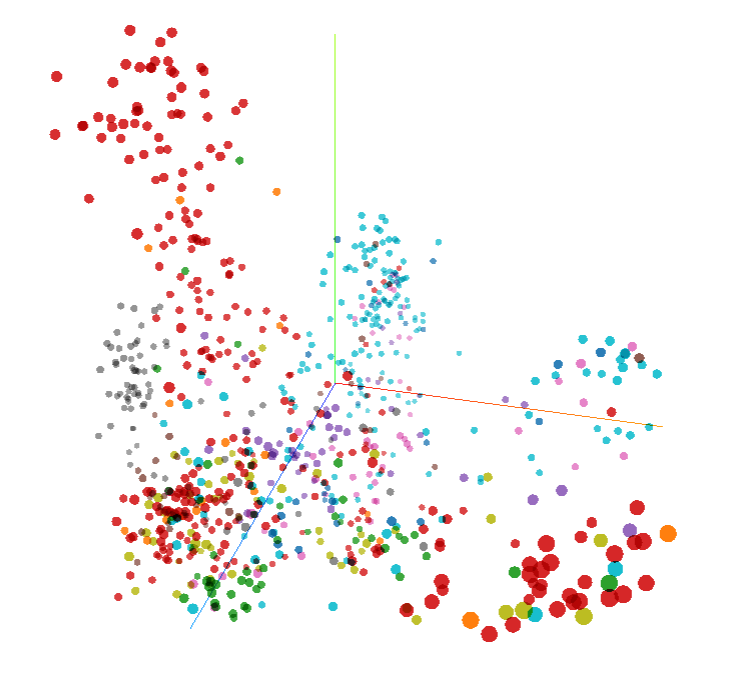}} &
\subcaptionbox{ALBEF-SS}{\includegraphics[width=0.45\linewidth]{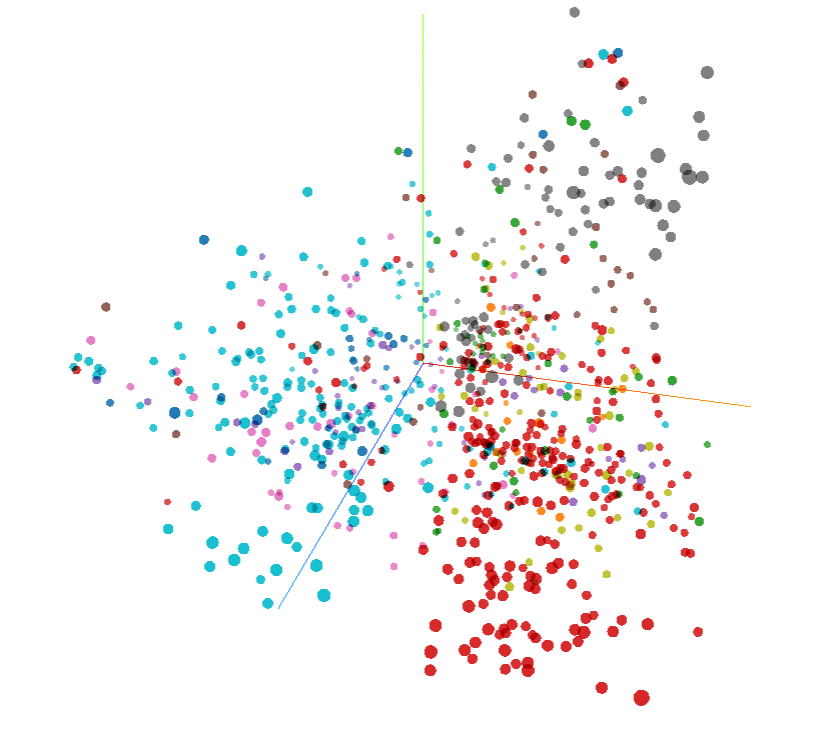}} 
\end{tabular}

 \caption{\footnotesize Joint embeddings from ALBEF and ALBEF-SS after PCA for top 20 groups (835 samples) in internal test set. }
\label{fig:test_data_dist}
\vspace{-5pt}
\end{figure*}

\subsubsection{Joint Embeddings from Internal test set:}
In figure 4, we show the PCA for joint embeddings of 20 most occurring groups in the internal test set (table 2 supp). As expected,  breast tissue density is the most common key concept in radiology reports.

%
% ---- Bibliography ----
%
% BibTeX users should specify bibliography style 'splncs04'.
% References will then be sorted and formatted in the correct style.
%
% \bibliographystyle{splncs04}
% \bibliography{references}

% \end{document}

\end{document}